\documentclass[journal,twoside]{IEEEtran}
\usepackage{epsfig}
\usepackage{graphicx}
\usepackage{subfigure}
\usepackage{multirow}
\usepackage{algorithm}
\usepackage{algorithmic}

\makeatletter
\newcommand{\Rmnum}[1]{\expandafter\@slowromancap\romannumeral #1@}

\ifCLASSINFOpdf

\else

\fi
\begin{document}
\title{A novel particle swarm optimizer with multi-stage transformation  and genetic operation for VLSI routing}

\author{Genggeng~Liu,
        Zhen~Zhuang,
        Wenzhong~Guo,
        Naixue Xiong,
        and~Guolong~Chen
\thanks{}
\thanks{}
}


\markboth{November~2018}%
{Liu \MakeLowercase{\textit{et al.}}: A novel particle swarm optimizer with multi-stage transformation  and genetic operation for VLSI routing}

\maketitle

\begin{abstract}
As the basic model for very large scale integration (VLSI) routing, the Steiner minimal tree (SMT) can be used in various practical problems, such as wire length optimization, congestion, and time delay estimation. In this paper, a novel particle swarm optimization (PSO) algorithm  based on multi-stage transformation and genetic operation is presented to construct two types of SMT, including non-Manhattan SMT and Manhattan SMT. Firstly, in order to be able to handle two types of SMT problems at the same time, an effective edge-vertex encoding strategy is proposed.  Secondly, a multi-stage transformation strategy is proposed to both expand the algorithm search space and ensure the effective convergence. We have tested three types from two to four stages and various combinations under each type to highlight the best combination. Thirdly, the genetic operators combined with union-find partition are designed to construct the discrete particle update formula for discrete VLSI routing. Moreover, in order to introduce uncertainty and diversity into the search of PSO algorithm, we propose an improved mutation operation with edge transformation. Experimental results show that our algorithm from a global perspective of multilayer structure can achieve the best solution quality among the existing algorithms. Finally, to our best knowledge, it is the first work to address both manhattan and non-manhattan routing at the same time.
\end{abstract}

\begin{IEEEkeywords}
Particle swarm optimization (PSO), multi-stage transformation, genetic operation, very large scale integration (VLSI), X-architecture routing, Manhattan routing.
\end{IEEEkeywords}

\IEEEpeerreviewmaketitle

\section{Introduction}
\IEEEPARstart{G}{lobal} routing is a very important step in very large scale integration (VLSI) physical design. The Steiner minimal tree (SMT) problem is to find a minimum cost routing tree that connects a given pin set by introducing additional points (Steiner points). The Steiner minimum tree is the best connection model for the multi terminal net in VLSI global routing and thus SMT construction is a key point .

\indent At present, most routing algorithms were proposed for rectilinear architecture [1-8]. With the development of VLSI technology, interconnect effect is becoming the main challenge of the performance of chip. However, the routing model based on rectilinear architecture requires that the connection between the chip pins can only be horizontal and vertical, resulting in limited ability to optimize the interconnect length in the chip. As a result, more and more researchers are interested in the non-Manhattan architecture which can make full use of routing resources and bring better length optimization capability [9-16].

\indent Constructing the non-Manhattan Steiner tree is a NP hard problem [17]. On one hand, the above researches on non-Manhattan Steiner tree [9-16] are both based on exact algorithm and traditional heuristic algorithm. However, the time complexity of the exact algorithm increases exponentially with the scale of the problem. Most of the traditional heuristic algorithms are based on greedy strategy and easy to fall into local minima. The two types of methods in the construction of Steiner tree, did not make full use of the geometric properties of non-Manhattan architecture, and cannot guarantee the quality of the Steiner tree. Those methods provided less suitable method for the topology optimization and therefore it is not satisfactory in terms of time efficiency, wirelength and so on.

On the other hand, as a swarm-based evolutionary method, particle swarm optimization (PSO) was introduced by Eberhart and Kennedy [18], which has been proved to be a global optimization algorithm. PSO algorithm has quick convergence, global search, stable, efficient, and many other advantages. Therefore, in recent years, more and more PSO algorithms are used to solve the NP hard problem and achieve good results [19-24].

For this reason, a novel particle swarm optimizer with multi-stage transformation and genetic operators, namely NPSO-MST-GO, is proposed to construct two type of SMT, including non-Manhattan SMT and Manhattan SMT. The main contributions of this paper can be summarized as follows.

\begin{itemize}
\item
An effective edge-vertex encoding strategy with four types of pseudo-Steiner point choice is proposed for the non-Manhattan Steiner tree. And when the proposed encoding strategy uses two types of of pseudo-Steiner point choice, it can be effectively extended to construct the Manhattan Steiner tree. Therefore, it helps the proposed algorithm can effectively two types of SMT.
\item
Meanwhile, a multi-stage transformation strategy is proposed to both expand the algorithm search space and ensure the effective convergence. We have tested three types from two to four stages and varous combinations under each type to highlight the best combination for the proposed algorithm. Therefore, it can construct SMT with less wirelength.
\item
In order to bring uncertainty and diversity into the proposed PSO algorithm, four genetic operators are proposed. Genetic operators also help the proposed algorithm to address the discrete VLSI routing.
\item
A series of simulation experiments are designed to illustrate the feasibility and effectiveness of the proposed strategies and can achieve better solution with less runtime.
\end{itemize}

 The remainder of the paper is organized as follows. Section \Rmnum{2} introduces the related work of VLSI routing problem. In Section \Rmnum{3}, the preliminaries are described. In Section \Rmnum{4}, the corresponding implementation details and the global convergence proof of NPSO-MST-GO algorithm are described. To prove the good performance of our proposed NPSO-MST-GO, several comparisons are presented and experimental results are discussed in Section \Rmnum{5}. Section \Rmnum{6} concludes this paper.

\section{Related work}
\indent In recent years, integrated circuits have been developing rapidly in both design scale and manufacturing process, thus it brings new challenges for electronic design automation. The global routing problem of physical design is becoming more and more complex. The research work of the global routing problem is mainly focused on the routing tree construction and the design of the global routing algorithm in rectilinear architecture, and some good research work has been obtained [1-4]. Especially in the VLSI global routing algorithm competition held in ISPD, NCTU-R [5], NTHU-Route2.0 [6], NCTU-GR [7], NCTU-GR2.0 [8] and other global routers  have emerged, and stand out in the global routing competition.

Manhattan architecture which restricts the routing directions to be only horizontal and vertical has limited the ability to optimize wire length and timing delay. Consequently, more and more researchers begin put their work into non-Manhattan architecture, which allows more routing
directions and could further improve the routability. Especially, the special industry alliance arises to promote X-architecture and it provides the foundation of implementation and verification for the study of X-architecture.

\indent In order to better study the routing problem based on non-Manhattan architecture, the first work is to study the construction of Steiner minimal tree in non-Manhattan architecture. The branch and bound method was proposed to construct a hexagonal Steiner tree, but only suitable for small scale problems [9]. Coulston used the accurate algorithm and a series of pruning strategies to construct the octagonal routing tree [10]. Compared with the rectilinear Steiner minimal tree (RSMT), it can possibly bring a 1.16\% reduction in wirelength, but it pays a higher cost of algorithm complexity. In view of the high time complexity of exact algorithms, people began to explore the application of heuristic strategies in constructing non-Manhattan architecture Steiner trees. An algorithm with the time complexity ${O(|V|+|E|)}$ was presented to construct the octagonal Steiner tree [11], but the constructed Steiner tree must be isomorphic. A heuristic algorithm with the time complexity $O(n^3logn)$ was presented to solve the hexagonal Steiner tree problem [12], which is based on greedy algorithm. Two algorithms based on octagonal spanning graph were proposed for the octagonal architecture Steiner tree construction [13]. One is the use of edge replacement technology. The other one is the triangle contraction method, which can obtain relatively good results with the slightly increase the time cost. The two methods are based on greedy algorithms. A modified algorithm is proposed to construct the time-driven octagonal Steiner tree. By adjusting the longest path of the hexagonal routing, the delay of the longest path is effectively reduced [14]. Based on Hanan grid and Elmore delay model, Samanta et al. constructed the timing driven Steiner tree, and verified that the Elmore delay model has higher computational accuracy and fidelity than the linear delay model [15]. Based on graph theory, $k$-IDEA algorithm was used to generate multiple sets of routing solutions [16]. The branch boundary concept was introduced to accelerate the solving procedure. The smaller wirelength and congestion is obtained, but excessive vias are generated.

The minimum Steiner tree problem is a NP hard problem, so some evolutionary algorithms which have shown a good application prospect in solving NP hard problems were used to solve RSMT [25-30] and X-architecture architecture Steiner minimum tree (XSMT) problem [29,31]. As a swarm-based evolutionary method, particle swarm optimization (PSO) was introduced by Eberhart and Kennedy [18], which has been proved to be a global optimization algorithm. PSO algorithm has quick convergence, global search, stable, efficient, and many other advantages [19-24]. In recent years, PSO algorithm has been widely and successfully used in many research fields, and has been proved to be a powerful optimization tool by means of continuous simulation and theoretical analysis [32-37].
\section{Preliminaries}
\subsection{Problem Formulation}
\textit{Definition 1 ($\lambda$-geometry):}
In the $\lambda$-geometry, the routing direction is $i\pi /\lambda $, where $i$ is an arbitrary integer and $\lambda$ is an integer. Different routing directions are obtained with different values of $i$ and $\lambda$.

\textit{1) Rectilinear architecture:} The value of $\lambda$ is set to 2, i.e. the routing direction is $i\pi /2$, which includes \emph{$0^{0}$} and \emph{$90^{0}$}, namely traditional horizontal and vertical orientations. Rectilinear architecture belongs to Manhattan architecture.\par

\textit{2)  X-architecture:} The value of $\lambda$ is set to 4¡A i.e. the routing direction is $i\pi /4$, which includes \emph{$0^{0}$}, \emph{$45^{0}$}, \emph{$90^{0}$} and \emph{$135^{0}$}.  X-architecture belongs to non-Manhattan architecture.
 \par

\indent The SMT problem based on X-architecture is more complicated and challenging than the one based on rectilinear architecture. Therefore, we first study the efficient algorithm for the XSMT problem, and then extend it to be applied in RSMT problem.\par

\textit{XSMT problem:} Given a set of $n$ pins \emph{P} = \{\emph{$P_{1}$}, \emph{$P_{2}$}, \emph{$P_{3}$}, \ldots, \emph{$ P_{m}$}\} and each \emph{$P_{i} $} represented by a coordinate of (\emph{$ x_{i} $}, \emph{$ y_{i} $}), construct an octagonal Steiner minimal tree to connect the pins in \emph{P} through some Steiner points, where the direction of routing path can be  \emph{$45^{0}$} and \emph{$135^{0}$}, in addition to the traditional horizontal and vertical directions. For example, there is a routing net with 8 pins and the input information of the pins is given in Table \Rmnum{1}. The layout distribution of the given pins is shown in Fig. 1. The coordinate pairs of pin 1 is (33,33), as shown in Column 2 of Table \Rmnum{1}.\par

\begin{table}[htbp]
  \centering
  \caption{The input information for the pins of net}
    \begin{tabular}{rrrrrrrrr}
    \hline
    Number    & 1     & 2     & 3     & 4     & \multicolumn{1}{c}{5} & 6     & 7     & 8 \\
    \hline
    X-coordinate & 33    & 2     & 42    & 47    & 34    & 38    & 37    & 20 \\
    Y-coordinate & 33    & 9     & 35    & 2     & 1     & 2     & 5     & 4 \\
    \hline
    \end{tabular}%
  \label{tab:addlabel}%
\end{table}%

\label{sec:2.1}
\begin{figure}[]
\begin{minipage}[] {\linewidth}
\centering
\includegraphics[width=60mm]{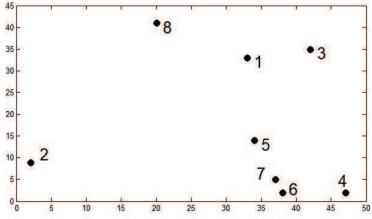}
\caption{The layout distribution of the given pins in Table 1}\label{fig:1}
\end{minipage}
\end{figure}
\begin{figure*}[]
\centering
\subfigure[]{
\label{fig:subfig:a} 
\includegraphics[width=30mm,height=33mm]{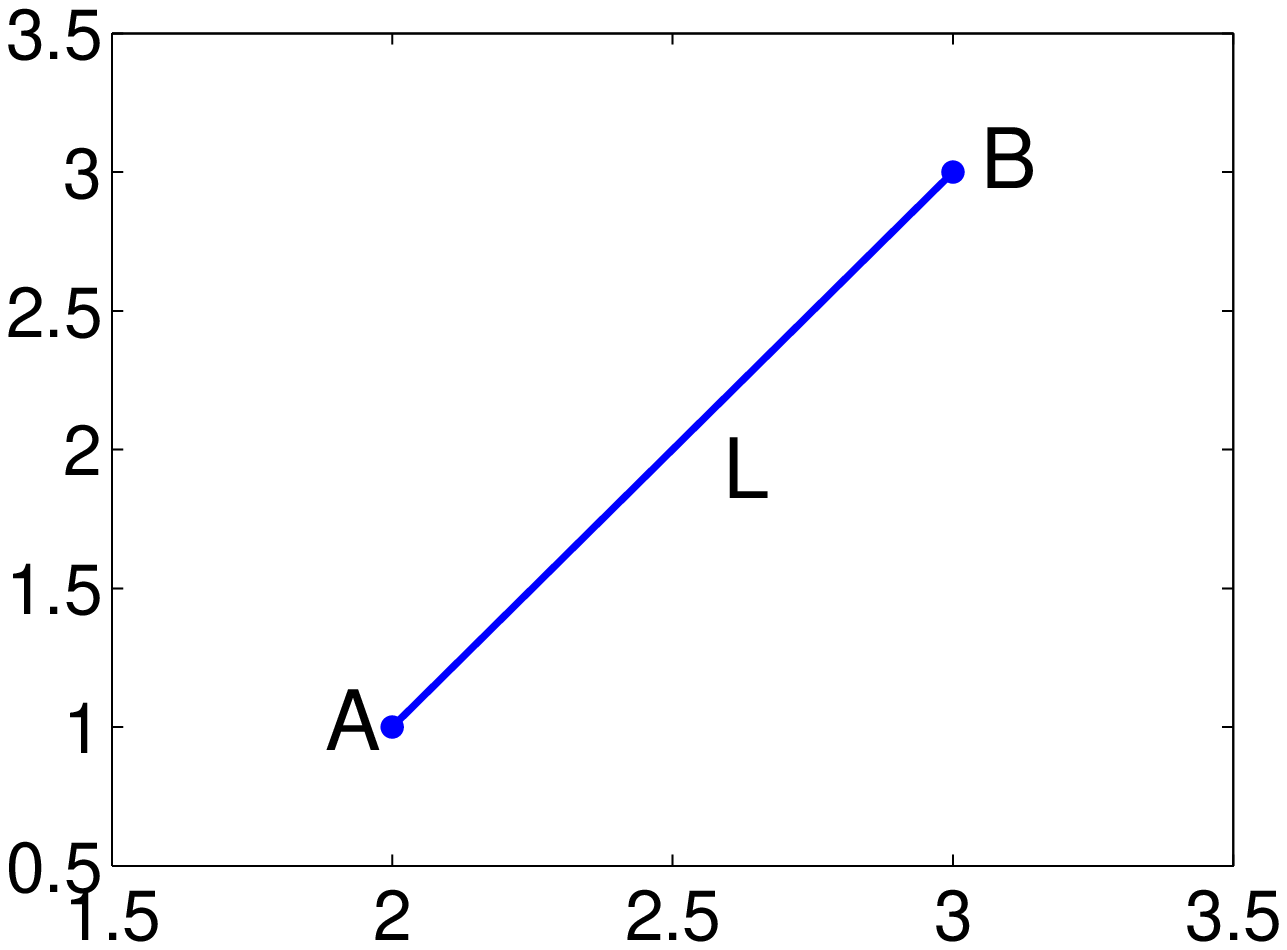}}
\subfigure[]{
\label{fig:subfig:b} 
\includegraphics[width=30mm,height=33mm]{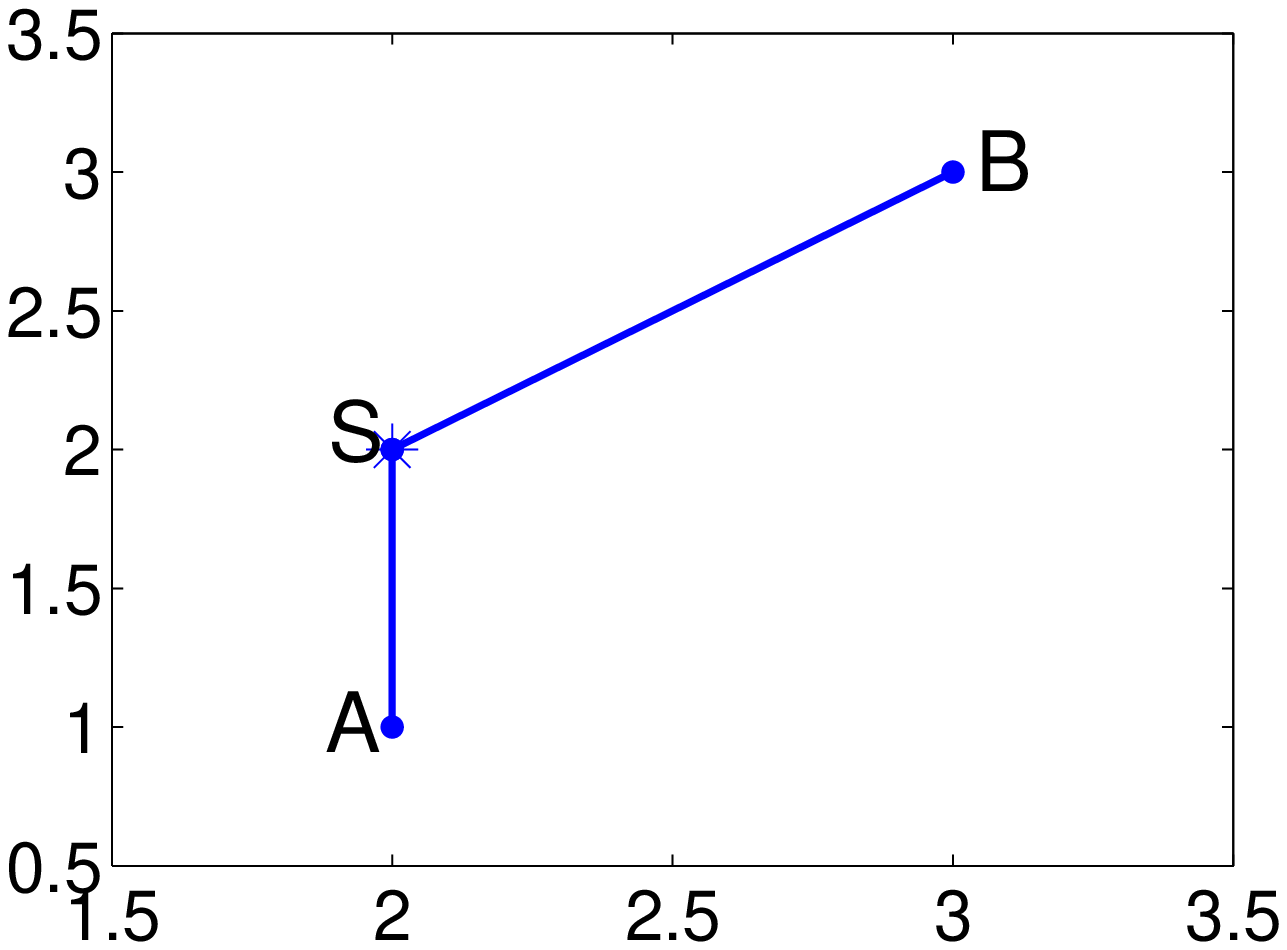}}
\subfigure[]{
\label{fig:subfig:c} 
\includegraphics[width=30mm,height=33mm]{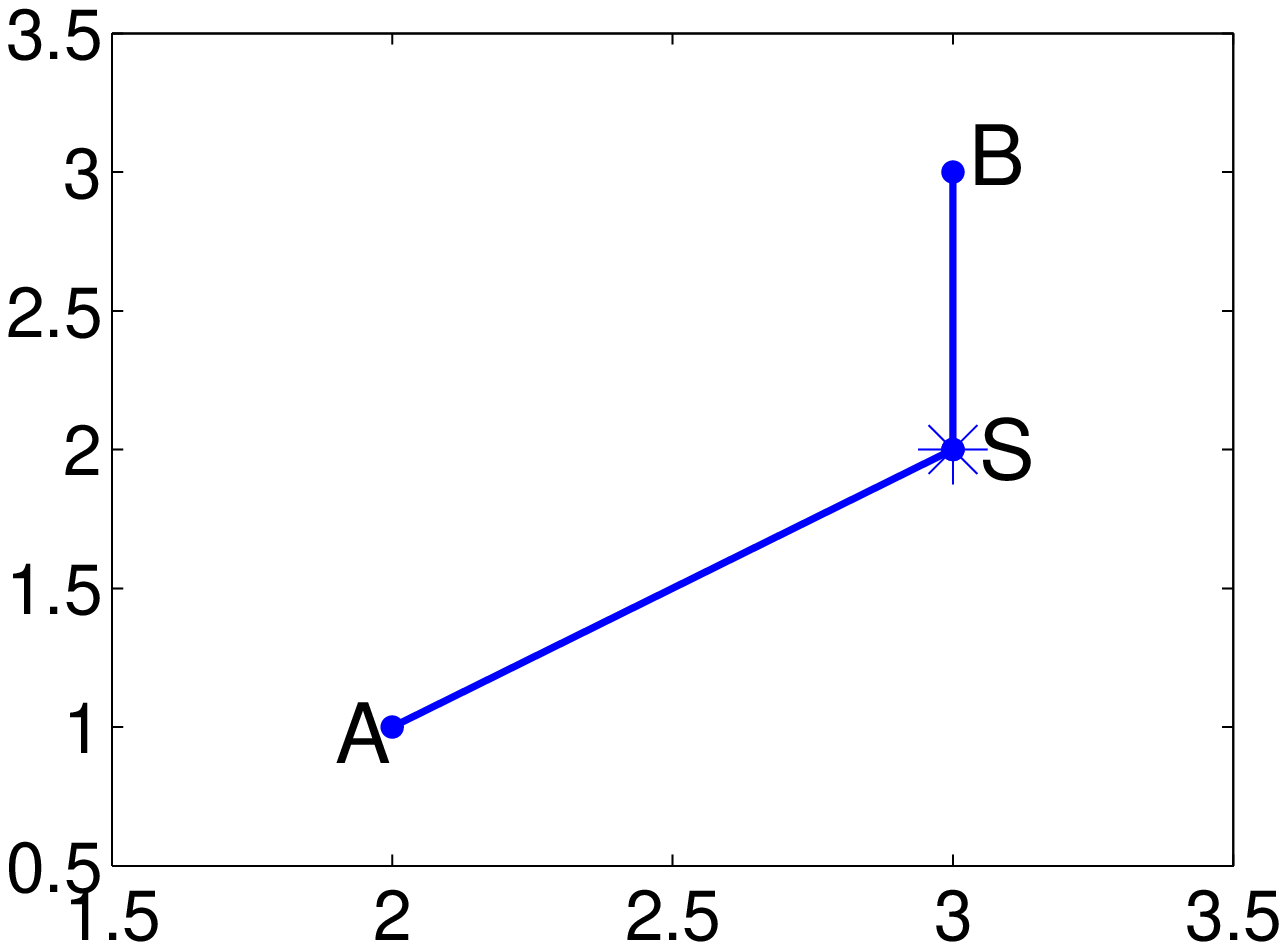}}
\subfigure[]{
\label{fig:subfig:d} 
\includegraphics[width=30mm,height=33mm]{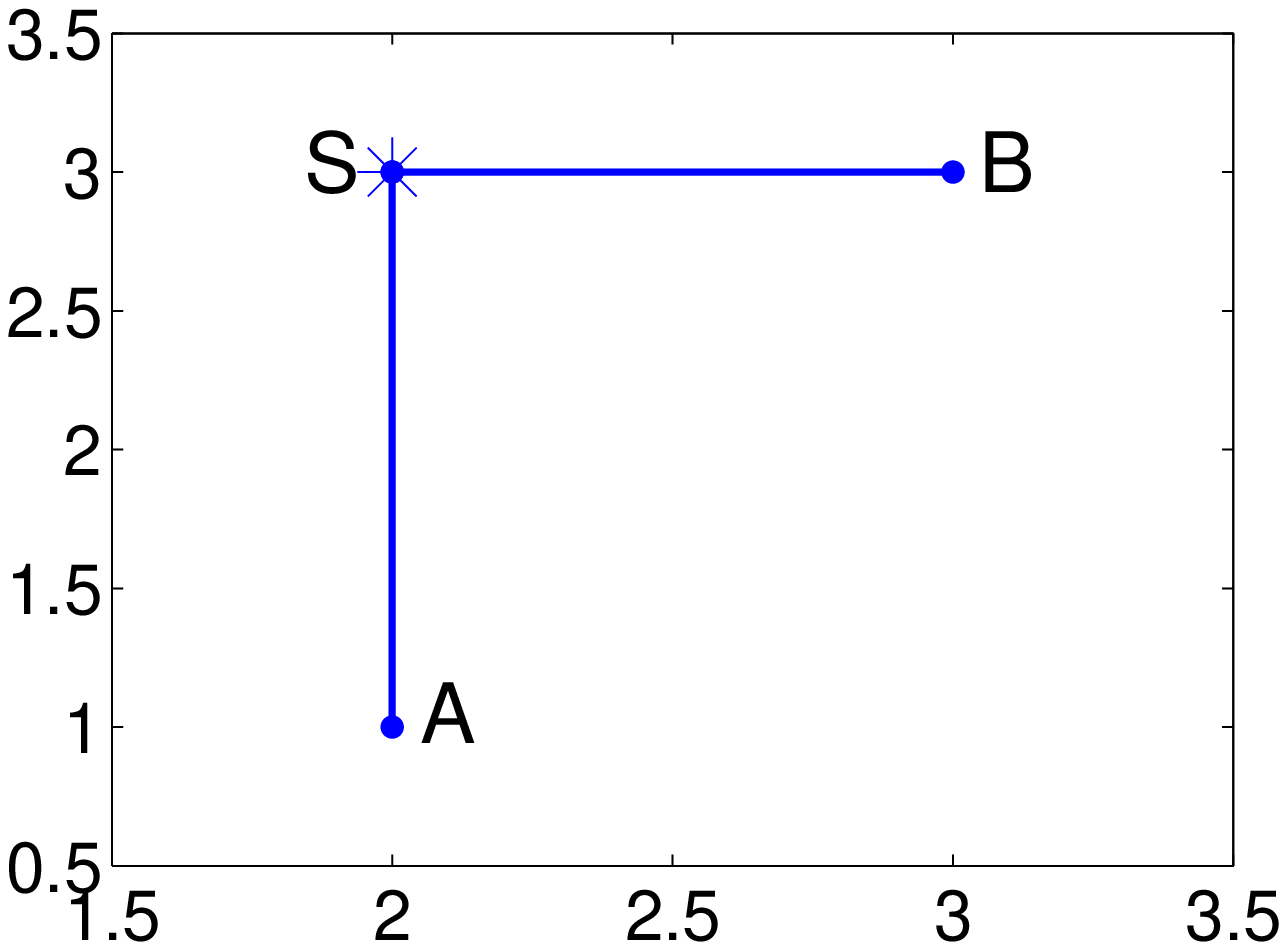}}
\subfigure[]{
\label{fig:subfig:e} 
\includegraphics[width=30mm,height=33mm]{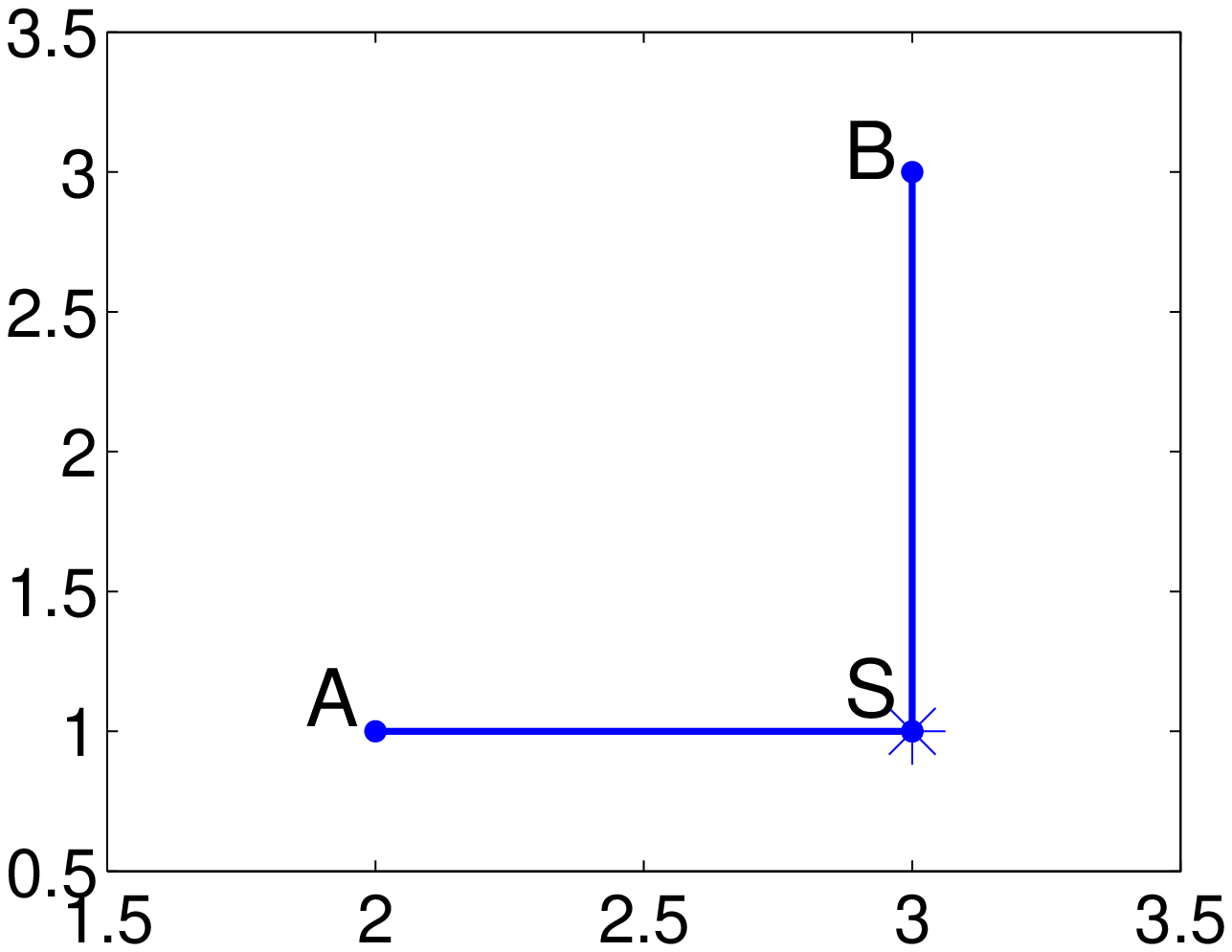}}
\caption{Four options of Steiner point for the given line segment. \textbf{a} Line segment \emph{L}, \textbf{b} 0 Choice, \textbf{c} 1 Choice, \textbf{d} 2 Choice, \textbf{e} 3 Choice}
\label{fig:2:subfig} 
\end{figure*}

\subsection{Definitions}
\label{sec:2.3}
\textit{Definition 2 (pseudo-Steiner Point)} For convenience, we assume that the connection point except for the pins is called the pseudo-Steiner point ('PS' represents 'the pseudo-Steiner point' in subsequent content). In Fig. 2, the point $S$ is PS, and PS contains the Steiner point.
\par
\textit{Definition 3 (0 Choice)} In Fig. 2(a), let \emph{A} = (\emph{x$_1$}, \emph{y$_1$}) and \emph{B} = (\emph{x$_2$}, \emph{y$_2$}) be the two endpoints of a line segment \emph{L}, where \emph{x$_1$ $<$ x$_2$}. The 0 Choice of \emph{L}  is given in Fig. 2(b), which first from \emph{A}  leads rectilinear side to pseudo-Steiner point $S$  and then leads octagonal side to \emph{B}.
\par
\textit{Definition 4 (1 Choice)} The 1 Choice of \emph{L}  is given in Fig. 2(b), which first from \emph{A}  leads octagonal side to pseudo-Steiner point $S$  and then leads rectilinear side to \emph{B}.
\par
\textit{Definition 5 (2 Choice)} The 2 Choice of \emph{L}  is given in Fig. 2(c), which first from \emph{A}  leads vertical side to pseudo-Steiner point $S$ and then leads horizontal side to \emph{B}.
\par
\textit{Definition 6 (3 Choice)} The 3 Choice of \emph{L}  is given in Fig. 2(d), which first from \emph{A} leads horizontal side to pseudo-Steiner point $S$  and then leads vertical side to \emph{B}.
\par

\section{Details of NPSO-MST-GO algorithm}
\label{sec:3}
\subsection{Basic PSO Algorithm}
\label{sec:2.3}
PSO is a swarm intelligence method, which considers a swarm containing \emph{p} particles in a \emph{D}-dimensional continuous solution space. Each \emph{i}-th particle has its own position and velocity. Assuming that the search space is \emph{D}-dimensional, the position of the \emph{i}-th particle is denoted as a \emph{D}-dimensional vector: \emph{X$_{i}$} = (\emph{X $_{i1}$}, \emph{X$_{i2}$}, \ldots, \emph{X$_{iD}$}) and the best particle in the swarm is denoted as $P_g$. The best previous position of the \emph{i}-th particle is recorded and represented as \emph{P$_i$} = (\emph{P$_{i1}$}, \emph{P$_{i2}$}, \ldots, \emph{P$_{iD}$}), while the velocity for the \emph{i}-th particle can be defined by another \emph{D}-dimensional vector: \emph{V$_i$} = (\emph{V$_{i1}$}, \emph{V$_{i2}$}, \ldots, \emph{V$_{iD}$}). According to these definitions, the particle position and velocity can be manipulated according to the following equations:\par
\begin{equation}
V_i^{t + 1} = w \times V_i^t + {c_1}{r_1}(P_{i}^t - X_i^t) + {c_2}{r_2}(P_{g}^t - X_i^t)
\end{equation}
\begin{equation}
X_i^{t + 1} = X_i^t + V_i^{t+1}
\end{equation}
\noindent where \emph{w} is the inertia weight; \emph{c$_1$} and \emph{c$_2$} are acceleration coefficients; \emph{r$_1$} and \emph{r$_2$} are both random numbers on the interval [0, 1).

From Eqs. (1) and (2), it is obvious that the basic PSO cannot obtain a discrete solution for the XSMT problem as a result of its continuous nature, thus there must be taken steps to the basic PSO. It has led to many discrete PSO algorithms for solving discrete problems. Three typical discrete PSO
algorithms have been proposed : (1) The speed of particle is viewed as the probability of position change. (2) The continuous PSO is directly discredited and used to solve the discrete problem. (3) The PSO operators are redefined according to the discrete problem. In
our previous works, the importance of VLSI physical design and the advantages of PSO have led to many research results in partitioning,
floorplanning and routing [31,38-41]. According to the XSMT problem, this paper presents an effective discrete PSO algorithm based on  multi-stage transformation and genetic operations. In Section \Rmnum{4}-B, we design a kind of edge-vertex encoding strategy which is suitable for
X-architecture and can be effectively extended to solve the RSMT problem.  Then, Section \Rmnum{4}-C proposes an effective fitness function to optimize the wirelength of routing tree. Two kinds of genetic algorithm operators are given in Section \Rmnum{4}-D and Section \Rmnum{4}-E, respectively. The details of multi-stage transformation are described in Section \Rmnum{4}-F. The parameter strategy and the step of NPSO-MST-GO algorithm are given in Section \Rmnum{4}-G and Section \Rmnum{4}-H, respectively. Finally, the analysis and convergence proof of NPSO-MST-GO algorithm is presented in Section \Rmnum{4}-I.

\subsection{Encoding Strategy}
The encoding strategy of a spanning tree can usually include two ways: Prufer number encoding [42] and edge-vertex encoding.

\textit{Definition 7 (Puffer number)}
 The Puffer number encoding gives a one-one mapping between the spanning tree of the $N$ nodes and the string of numbers of $n$-2. It is customarily used to mark these nodes with numbers. The resulting string is called the Puffer number.

\textit{Property 1}
 An edge-vertex encoding strategy is more suitable for evolutionary algorithms than Pruffer number. Furthermore, an efficient decoding of Pruffer number has time that is $O(nlogn)$, while edge-vertex encoding strategy does not have to spend time performing it.

The length of the Puffer number encoding is 1/3 of the length of the edge-vertex encoding. However, since one digit of the Puffer number encoding is changed, the topology of the entire spanning tree is greatly changed. And thus the optimal topological information of the particle cannot be well retained. However, edge-vertex encoding is more suitable for the PSO algorithm and retains the partial optimal topological information of particles during the iterative process. Meanwhile, changing one digit of the edge-vertex encoding does not have a huge impact on the topology of the spanning tree. Therefore, the NPSO-MST-GO algorithm adopts the edge-vertex encoding to encode a particle, which is more suitable for particle swarm optimization, genetic algorithm and other evolutionary algorithms.

One candidate Steiner tree is represented as lists of spanning tree edges and each edge augments with a pseudo-Steiner point choice which specifies the transformation from the spanning tree edge to the octagonal edge. Each pseudo-Steiner point choice, includes four types as shown in Definition 3-6 and the value is 0, 1, 2 or 3 which denotes 0 Choice, 1 Choice, 2 Choice or 3 Choice, respectively.

If a net has $n$ pins, a spanning tree would have $n$-1 edges and one extra digit which is the particle¡¦s fitness. Besides, two digits represent the two vertices of each edge, so the length of one particle is 3($n$-1) +1. For example, one routing tree ($n$=8) can be expressed as one particle whose code can be represented as the following numeric string.

7 6 $0$ 6 4 $1$ 7 5 $1$ 5 1 $2$ 1 3 $0$ 1 8 $1$ 5 2 $2$ \textbf{10.0100}

\noindent where the bold number '10.0100' is the fitness of the particle and each italic number represents the PS choice for each edge. The first substring (7,6,0) represents one edge of the spanning tree which composed of pin vertex 7 and pin vertex 6 and the pseudo-Steiner point choice which is 0 Choice in Definition 3.

\textit{Property 2}
 An edge-vertex encoding strategy with four types of pseudo-Steiner point choice is more effective for the XSMT problem than the one with two types of pseudo-Steiner point choice.

\begin{figure}[t]
\centering
\begin{minipage}[t] {\linewidth}
\subfigure[]{
\label{fig:subfig:a} 
\includegraphics[width=35mm,height=35mm]{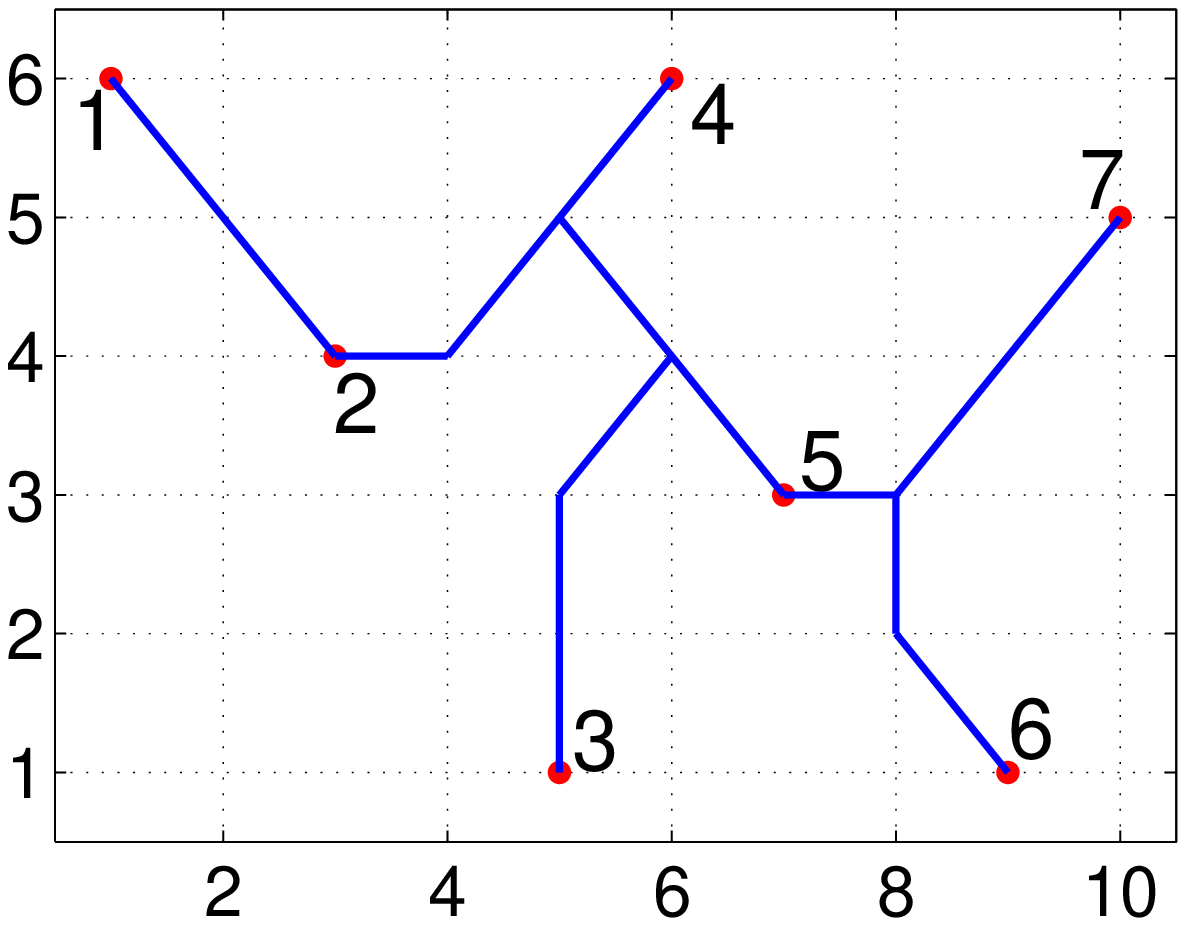}}
\subfigure[]{
\label{fig:subfig:b} 
\includegraphics[width=35mm,height=35mm]{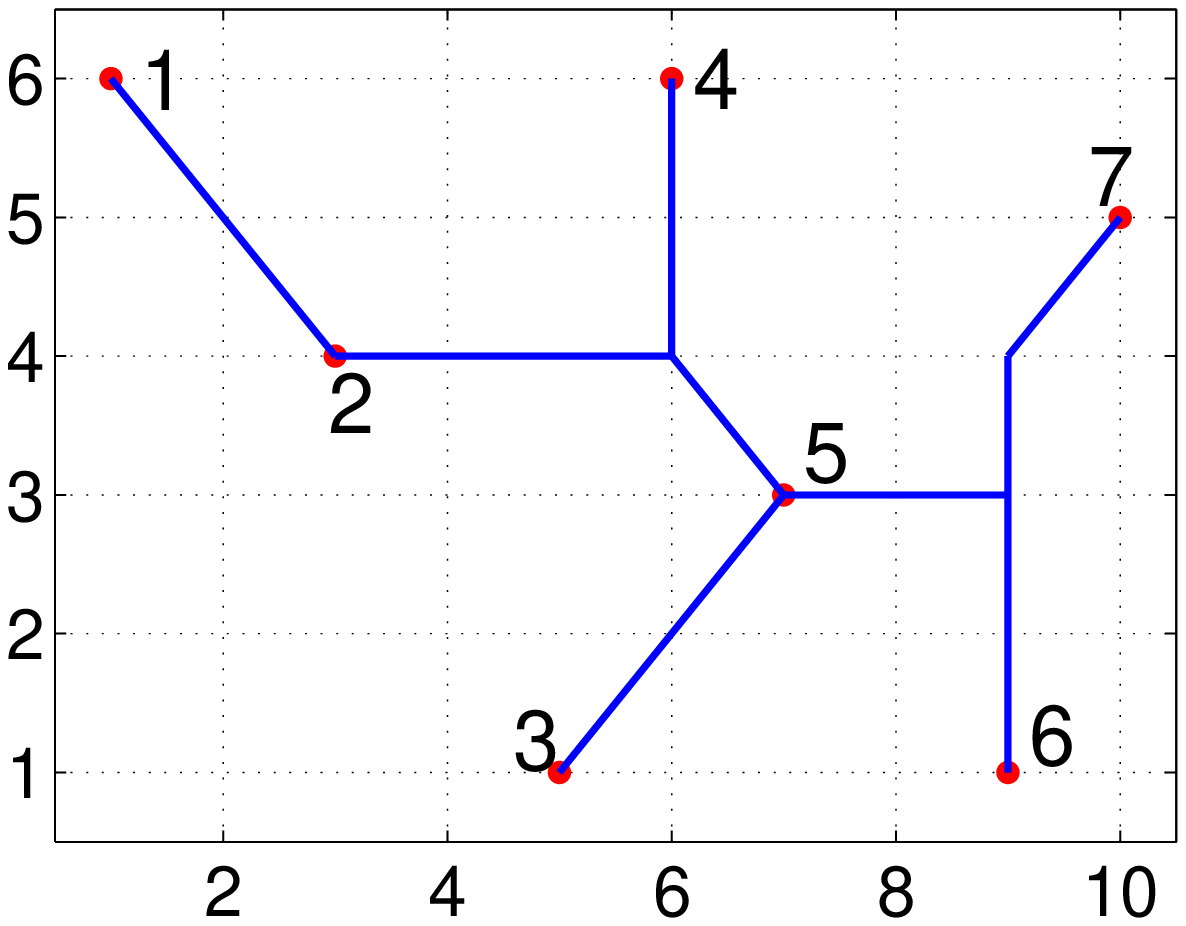}}
\caption{Two kinds of encoding strategies for the potential impact on the wirelength reduction. \textbf{a} The encoding strategy with two pseudo-Steiner point choices, \textbf{b} The encoding strategy with four pseudo-Steiner point choices}
\label{fig3:subfig} 
\end{minipage}%
\end{figure}

\indent The encoding strategy with two pseudo-Steiner point choices, i.e. including 0 Choice and 1 Choice shown in Fig. 2b and Fig. 2c. In contrast, the encoding strategy with four pseudo-Steiner point choices is applied in our proposed algorithm which has the potential ability to reduce the wirelength. The reason is that the existence of the last two pseudo-Steiner point choices maybe overlaps with the first two choices with less wirelength. For example, the SMT in Fig. 3b, have the more overlapping location than the one in Fig. 3a. And thus the wirelength (15.070) of SMT in Fig. 3b is less than the wirelength (19.140) of SMT in Fig. 3a. Generally, this situation could frequently exist among the net with more pins. Therefore, for the most part, using the second encoding scheme combined with edge transformation, as shown in Fig. 3b can be more helpful to reduce the wirelength than the former, as shown in Fig. 3a. From Table \Rmnum{2}, it can find that the encoding strategy with four pseudo-Steiner point choices can achieve 1.04\% wirelength reduction. Meanwhile, for the circuits with large scale, it can reduce more wirelength with the proposed encoding strategy.
\begin{table}[htbp]
  \centering
  \caption{Comparison between the encoding strategy with four pseudo-Steiner point choices and the one with two choices }
    \begin{tabular}{ccccc}
    \hline
    \multirow{2}[2]{*}{TEST} & \multirow{2}[2]{*}{PIN\#} & \multicolumn{1}{c}{\multirow{2}[2]{*}{Two choices}} & \multicolumn{1}{c}{\multirow{2}[2]{*}{Four choices}} & \multirow{2}[2]{*}{Imp} \\
          &       &       &       &  \\
    \hline
    1     & 8     & \multicolumn{1}{c}{16951} & \multicolumn{1}{c}{16900} & 0.30\% \\
    2     & 9     & \multicolumn{1}{c}{18041} & \multicolumn{1}{c}{18023} & 0.10\% \\
    3     & 10    & \multicolumn{1}{c}{19435} & \multicolumn{1}{c}{19397} & 0.20\% \\
    4     & 20    & \multicolumn{1}{c}{32218} & \multicolumn{1}{c}{32038} & 0.56\% \\
    5     & 50    & \multicolumn{1}{c}{48435} & \multicolumn{1}{c}{47888} & 1.13\% \\
    6     & 70    & \multicolumn{1}{c}{56444} & \multicolumn{1}{c}{55771} & 1.19\% \\
    7     & 100   & \multicolumn{1}{c}{69268} & \multicolumn{1}{c}{68157} & 1.60\% \\
    8     & 410   & \multicolumn{1}{c}{142157} & \multicolumn{1}{c}{139505} & 1.87\% \\
    9     & 500   & \multicolumn{1}{c}{155040} & \multicolumn{1}{c}{152314} & 1.76\% \\
    10    & 1000  & \multicolumn{1}{c}{220998} & \multicolumn{1}{c}{217171} & 1.73\% \\
    Average &       &       &       & 1.04\% \\
    \hline
    \end{tabular}%
  \label{tab:addlabel}%
\end{table}%

\textit{Property 3}
 An edge-vertex encoding strategy for the XSMT problem can be can be effectively extended to solve the RSMT problem.

If edge-vertex encoding strategy for the XSMT problem is designed to only contain two kinds pseudo-Steiner point choices, i.e. 2 Choice and 3 Choice, then the proposed algorithm can effectively solve the RSMT problem. Therefore, the proposed encoding strategy is helpful for the proposed NPSO-MST-GO algorithm to be effectively extended to solve the RSMT problem. And this situation has been experimentally verified in latter section.

\subsection{Fitness Function}
\textit{Definition 8}
The length of the Octilinear Steiner tree is the sum of the lengths of all the edge fragments in the routing tree, which is calculated as follows.
\begin{equation}
L({T_X}) = \sum\limits_{{e_i} \in {T_x}} {l({e_i}} )
\end{equation}
\noindent where \emph{l}(\emph{e$_i$}) represents the length of each segment \emph{e$_i$} in the tree \emph{T$_x$}.
\par
When calculating the sum of the length of each edge fragment in octagonal Steiner tree, all the edge fragments are divided into the following four types: horizontal edge segments, vertical edge segments, \emph{$45^{0}$} edge segments, and \emph{$135^{0}$} edge segments. The algorithm first divides all the edge fragments into the above four types, and then rotates the  \emph{$45^{0}$} edge clockwise to the horizontal edge, and simultaneously rotates the  \emph{$135^{0}$} edge fragment clockwise direction to the vertical edge. The horizontal edge is arranged from the bottom to top and then from left to right according to the size of the left pin. Simultaneously arrange the vertical edge from left to right and from bottom to top according to the size of the lower pin. Finally, the length of octagonal Steiner tree is the sum of the total length of these edges excluding repeated segments.
\par
The fitness value of the particle in PSO is usually smaller, the better the particle is represented. So the particle fitness function of the algorithm is designed as follows.
\begin{equation}
fitness=\frac{1}{L(T_{X})+1}
\end{equation}
\label{sec:3.2.5}

The SMT problem is the discrete optimization problem. Therefore, the particle update formula of basic PSO can be no longer suitable for solving the XSMT problem. For this reason, the crossover and mutation operators combined with union-find partition are designed to construct the discrete particle update formula for NPSO-MST-GO algorithm. The improved particle update formula is defined as follows.
\begin{equation}
X_i^t = {N_3}({N_2}({N_1}(X_i^{t - 1},w),{c_1}),{c_2})
\end{equation}
\noindent where \emph{w} is an inertia weight, \emph{c$_1$} and \emph{c$_2$} are acceleration constants. \emph{N$_1$} denotes the mutation operation and \emph{N$_2$}, \emph{N$_3$} denote the crossover operations. We assume that \emph{r$_1$}, \emph{r$_2$}, \emph{r$_3$} are random numbers on the interval [0, 1).\par

\subsection{Mutation Operator}
For different problems, there are many ways of mutation operation. There are two kinds of mutation operators in our previous work, including the PS transformation strategy [31] in Fig. 4 and the E transformation strategy in Fig. 5.

\indent  For the XSMT construction [31],it proposed the mutation operator with the PS transformation strategy and each PS selection method contains four choices, namely, 0 choices, 1 choices, 2 choices and 3 choices. Based on the PS transformation strategy, we find that the proposed algorithm takes the wirelength as the optimization target, and obtains the better wirelength optimization ratio relative to RSMT, and the average wirelength can be reduced by 9.68\%, as shown in Table \Rmnum{3}.

\textit{Property 4}
If the mutation operator with PS transformation strategy in Fig. 4 is designed, the optimization ability of XSMT construction algorithm is limited.
\par
\indent  If the algorithm uses the mutation operator with the PS transformation strategy, then it is a method based on the minimum spanning tree to construct the SMT. According to [43], in the rectilinear routing architecture, the wirelength of SMT based on the minimum spanning tree construction may be 1.5 times the exact solution, and has a certain distance from the optimal solution relative to the accurate RSMT. As shown in (6), $COST(MST)$ represents the length of the SMT based on the minimal spanning tree and the $COST(RSMT)$ represents the exact algorithm of the RSMT. For the octagonal routing problem with more routing direction, this maximum distance will be greater.
\begin{equation}
COST(MST)/COST(RSMT)<3/2
\end{equation}
\par
\indent  Therefore, in process of constructing SMT, only if the PS transformation is executed, the solution space does not necessarily contain the optimal solution and even 0.5 times farther from the optimal solution. Therefore, we need to redesign the transformation strategy. Then another basic transformation strategy, namely edge transformation.\par

The introduction of edge transformation further optimizes the wire length of the Steiner tree. However, the introduction of the edge transformation may lead to the appearance of loop and generate the invalid solution in the iterative process, which destroys the soundness principle of the particle encoding. How to overcome the shortcoming caused by the edge transformation in the evolutionary process? The union-find partition which keeps track of the components connected so far is integrated into the following update operators.\par
\begin{table}[htbp]
  \centering
  \caption{Comparison between PS transformation and RSMT}
    \begin{tabular}{rrrrr}
    \hline
    \multirow{2}[2]{*}{TEST} & \multirow{2}[2]{*}{PIN\#} & \multirow{2}[2]{*}{RSMT} & \multirow{2}[2]{*}{PS} & Imp(\%)/ \\
          &       &       &       & RSMT \\
    \hline
    1     & 8     & 17928 & 16918 & 5.64\% \\
    2     & 9     & 20478 & 18041 & 11.90\% \\
    3     & 10    & 21969 & 19696 & 10.35\% \\
    4     & 20    & 35675 & 32207 & 9.72\% \\
    5     & 50    & 53518 & 48020 & 10.27\% \\
    6     & 70    & 62633 & 56433 & 9.90\% \\
    7     & 100   & 75584 & 68855 & 8.90\% \\
    8     & 410   & 158206 & 141894 & 10.31\% \\
    9     & 500   & 170924 & 154825 & 9.42\% \\
    10    & 1000  & 246731 & 221090 & 10.39\% \\
    Average   &       &       &       & 9.68\% \\
    \hline
    \end{tabular}%
  \label{tab:addlabel}%
\end{table}%
\label{sec:3.2}
\textit{Property 5}
 The introduction of E transformation in the mutation operator, as shown in Fig. 5, extends the search space of the algorithm and it has the opportunity to obtain better routing solution than PS transformation. Even in some cases, it can help the algorithm to find the optimal solution.

 \begin{figure}[!ht]
\centering
\includegraphics[height=22mm]{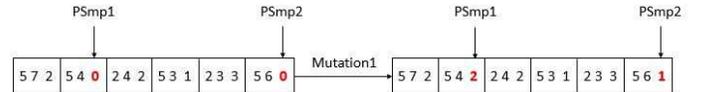}
\caption{Mutation operator with PS transformation.}\label{Fig.5}
\end{figure}

\begin{figure}[!ht]
\centering
\includegraphics[height=22mm]{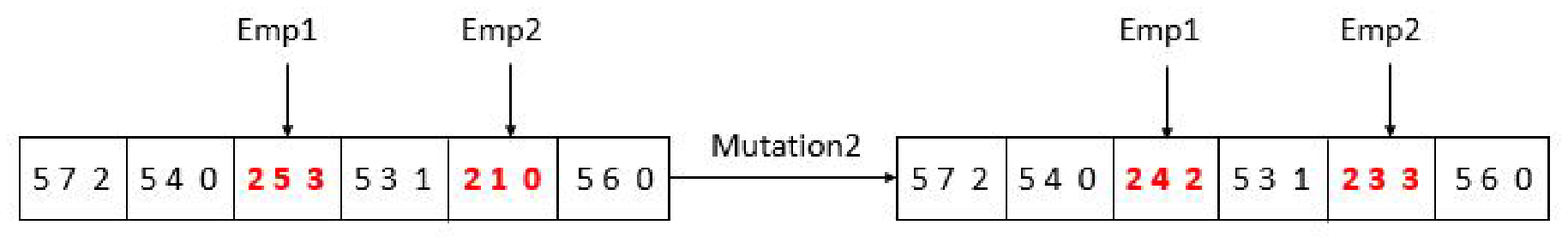}
\caption{Mutation operator with E transformation.}\label{Fig.5}
\end{figure}

\begin{table}[htbp]
  \centering
  \caption{Comparison among E transformation, PS transformation and RSMT}
    \begin{tabular}{rrrrrr}
    \hline
    \multirow{2}[2]{*}{TEST} & \multirow{2}[2]{*}{RSMT} & \multirow{2}[2]{*}{PS} & \multicolumn{1}{c}{\multirow{2}[2]{*}{E}} & Imp/  & Imp/ \\
          &       &       &       & RSMT  & PS \\
    \hline
    1     & 17928 & 16918 & 16921 & 5.62\% & -0.02\% \\
    2     & 20478 & 18041 & 18023 & 11.99\% & 0.10\% \\
    3     & 21969 & 19696 & 19397 & 11.71\% & 1.52\% \\
    4     & 35675 & 32207 & 32163 & 9.85\% & 0.14\% \\
    5     & 53518 & 48020 & 48027 & 10.26\% & -0.02\% \\
    6     & 62633 & 56433 & 56551 & 9.71\% & -0.21\% \\
    7     & 75584 & 68855 & 68991 & 8.72\% & -0.20\% \\
    8     & 158206 & 141894 & 141443 & 10.60\% & 0.32\% \\
    9     & 170924 & 154825 & 155820 & 8.84\% & -0.64\% \\
    10    & 246731 & 221090 & 221324 & 10.30\% & -0.11\% \\
    Average   &       &       &       & 9.76\% & 0.09\% \\
    \hline
    \end{tabular}%
  \label{tab:addlabel}%
\end{table}%

\begin{figure}[t]
\centering
\begin{minipage}[t] {\linewidth}
\subfigure[]{
\label{fig:subfig:a} 
\includegraphics[width=35mm,height=35mm]{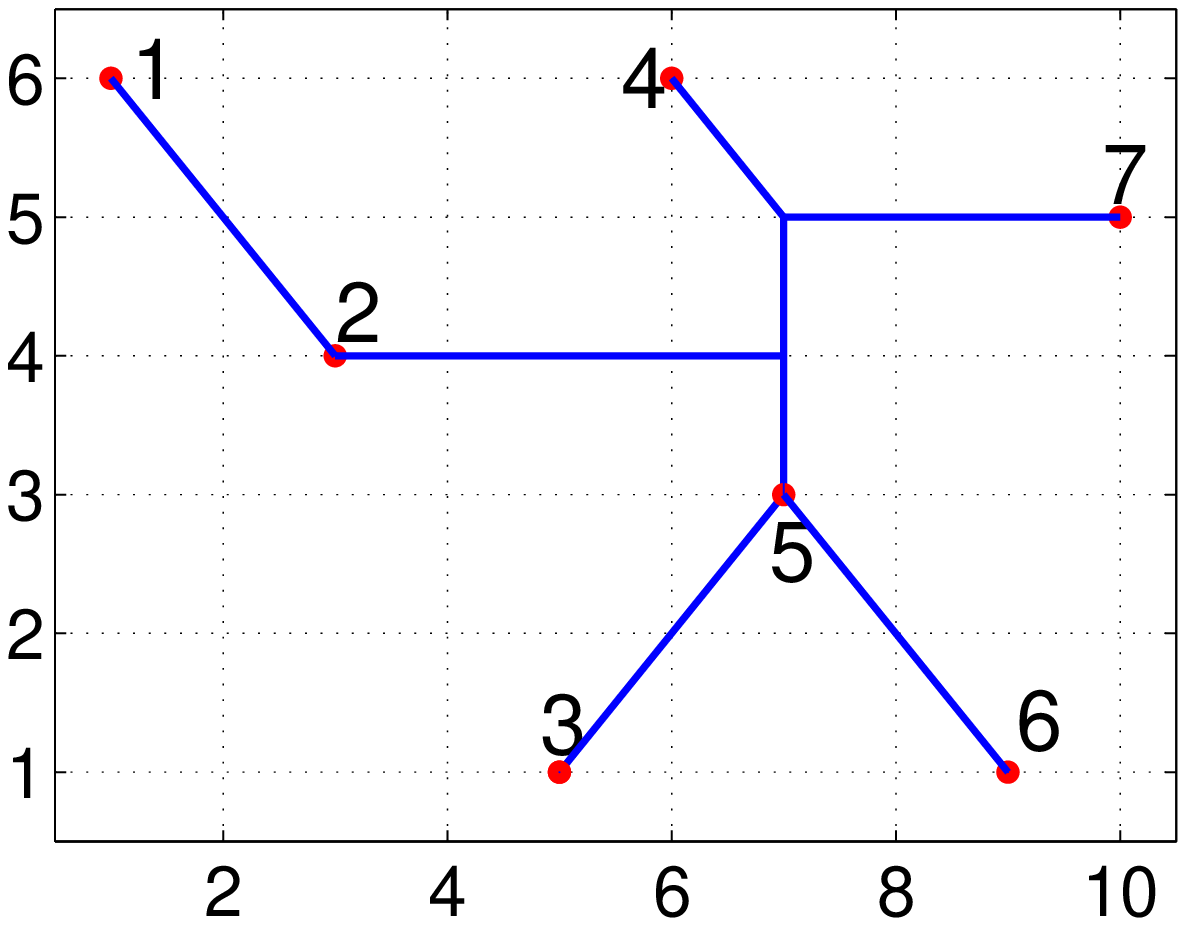}}
\subfigure[]{
\label{fig:subfig:b} 
\includegraphics[width=35mm,height=35mm]{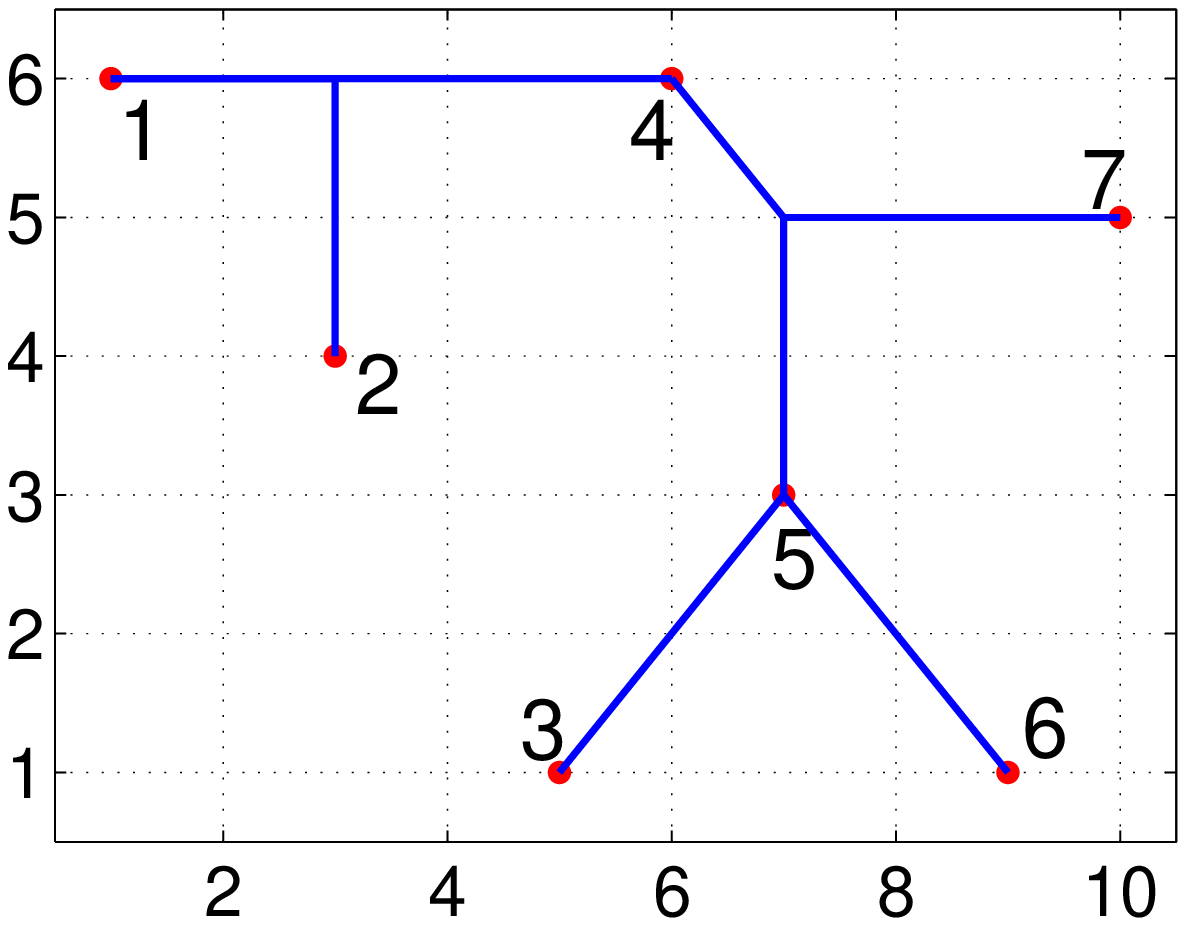}}
\caption{The necessity of E transformation in the evolutionary process of the algorithm. \textbf{a} Routing tree topology A, \textbf{b} Routing tree topology B}
\label{fig3:subfig} 
\end{minipage}%
\end{figure}

\textit{Property 6}
 With the introduction of E transformation, the search space of our proposed algorithm contains the exact solution for rectilinear routing.
\par
\indent  The update operation of PSO for constructing XSMT [31] only considers the PS transformation. However, in the process of building RSMT or XSMT, the topology of the routing tree is changed. For example, as shown in Fig.6, in the evolutionary process of constructing XSMT, there are two different topologies of routing tree. The edge (4,2) exists in the routing tree shown in  Fig.6b, but does not exist in the routing tree shown in  Fig.6a. In  Fig.6, the transformation of these two edges is not possible only through PS transformation strategy. Therefore, in the update operation of the RSMT or XSMT construction algorithm, E transformation strategy is needed in addition to PS transformation strategy. Moreover, E transformation strategy can further enhance the wirelength optimization ability. As shown in Table \Rmnum{4}, E transformation can reduce the wirelength by 0.09\% compared to PS transformation. Meanwhile, E transformation strategy is introduced into the update operation and it can ensure that the search space of the proposed algorithm [30] contains the optimal solution. And compared with the RSMT construction algorithm [30], the average wirelength reduction of 9.76\% is obtained.

\textit{Property 7}
It is not the more variation that happens, the better the solution it gets. Then through a lot of experiments, it can find that two point mutation operator is more effective for the proposed algorithm than others.

In order to find more effective mutation operator for our algorithm, we also test the single point mutation operator, two point mutation operator, three point mutation operator, and four point mutation operator. From Table \Rmnum{5}, we can find that the two point mutation operator achieve the best wirelength. Therefore, the two point mutation operator is adopted in the proposed algorithm.

\begin{table}[htbp]
  \centering
  \caption{Comparison among four mutation operators}
    \begin{tabular}{ccrcccc}
    \hline
    \multirow{2}[2]{*}{TEST} & \multirow{2}[2]{*}{PIN\#} & \multicolumn{1}{c}{\multirow{2}[2]{*}{RMST}} & \multirow{2}[2]{*}{1-Mutation} & \multirow{2}[2]{*}{2-Mutation} & \multirow{2}[2]{*}{3-Mutation} & \multirow{2}[2]{*}{4-Mutation} \\
          &       &       &       &       &       &  \\
    \hline
    1     & 8     & \multicolumn{1}{c}{17928} & 5.70\% & 5.74\% & 5.74\% & 5.70\% \\
    2     & 9     & \multicolumn{1}{c}{20478} & 11.99\% & 11.99\% & 11.99\% & 11.99\% \\
    3     & 10    & \multicolumn{1}{c}{21969} & 11.71\% & 11.71\% & 11.71\% & 11.71\% \\
    4     & 20    & \multicolumn{1}{c}{35675} & 10.04\% & 10.20\% & 10.13\% & 10.09\% \\
    5     & 50    & \multicolumn{1}{c}{53518} & 10.44\% & 10.52\% & 10.43\% & 10.49\% \\
    6     & 70    & \multicolumn{1}{c}{62633} & 10.76\% & 10.96\% & 10.48\% & 10.58\% \\
    7     & 100   & \multicolumn{1}{c}{75584} & 9.66\% & 9.83\% & 9.77\% & 9.75\% \\
    8     & 410   & \multicolumn{1}{c}{158206} & 11.61\% & 11.73\% & 11.65\% & 11.53\% \\
    9     & 500   & \multicolumn{1}{c}{170924} & 10.87\% & 10.89\% & 10.87\% & 10.85\% \\
    10    & 1000  & \multicolumn{1}{c}{246731} & 11.93\% & 11.98\% & 11.93\% & 11.84\% \\
    Average &       &       & 10.47\% & 10.55\% & 10.47\% & 10.45\% \\
    \hline
    \end{tabular}%
  \label{tab:addlabel}%
\end{table}%

With the proposed best mutation operator, the velocity of particles can be written as follows.
\begin{equation}
W_i^t = {N_1}(X_i^{t - 1},w) = \left\{ \begin{array}{l}
M(X_i^{t - 1}),{\rm{   }}{r_1} < w\\
X_i^{t - 1},{\rm{         others}}
\end{array} \right.
\end{equation}
\noindent where \emph{w} denotes the mutation probability. \par
Mutation operator randomly deletes an edge from the spanning tree and replaces it with another randomly generated edge. In order to make sure that the spanning tree is connected, we use a union-find partition to record all the rest of points as two sets after one edge is deleted, and randomly select a point from the two point sets respectively, and then connect the two points to form a new spanning tree. The pseudo code of mutation operator is shown in Algorithm 1.

\begin{algorithm}[t]
\caption{Mutation operator($p$)}
\begin{algorithmic}
\REQUIRE Particle  $p$ \ENSURE New particle \STATE $Initialize ~each
~pin's ~partition ~to ~singletons$ \STATE $r=random(1,n-1);$ //$n$
is the number of pins \FOR{$each~edge~e_i ~of ~p$} \IF {$e_i \neq
e_r$} \STATE $Union\_partition(u,v);$//$u$ and $v$ is endpoint of
$e_i$, $u$ and $v$ are merged into the same set \ENDIF \ENDFOR
\WHILE{$true$} \STATE$p_1= random(1,n-1);$// generate a random
number between (1,$n$-1) \STATE$p_2= random(1,n-1);$ \IF
{$Find\_set(p_1)\neq Find\_set(p_2)$}  \STATE
$Union\_partition(p_1,p_2);$//$p_1$ and $ p_2$ are not in the same
set \STATE $generate\_edge(p_1,p_2);$ \STATE $break;$ \ENDIF
\ENDWHILE
\end{algorithmic}
\end{algorithm}
\subsection{Crossover Operator}
Similar to the mutation operator, there are two kinds of crossover operator, shown in Fig. 7 and Fig 8, respectively. In Fig. 7, the  intersection point is only included in the PS point, while the intersection point is included in both the PS point and the edge location in Fig. 8.

With the introduction of crossover operator, the cognitive personal experience of particles can be written as follows.
\begin{equation}
S_i^t = {N_2}(W_i^t,{c_1}) = \left\{ \begin{array}{l}
{C_p}(W_i^t),{\rm{   }}{r_2} < {c_1}\\
W_i^t,{\rm{         others}}
\end{array} \right.
\end{equation}
\noindent where  \emph{c$_1$}  denotes the crossover probability between the particles and the personal optimal solution.\par
With the introduction of crossover operator, the cooperative global experience of particles can be written as follows.
\begin{equation}
X_i^t = {N_3}(S_i^t,{c_2}) = \left\{ \begin{array}{l}
{C_g}(S_i^t),{\rm{   }}{r_3} < {c_2}\\
S_i^t,{\rm{         others}}
\end{array} \right.
\end{equation}
\noindent where  \emph{c$_2$}  denotes the crossover probability between the particles and the global optimal solution.\par
When implementing the crossover operator in (8) and (9), we sort the edges according to the serial number of pins from small to large order in both of the two spanning tree, then by using union-find partition to choose the same edge from the two sorted tree as a set, the rest of the different edges as another set, and the first set of edges directly as the new spanning tree¡¦s edges. Finally, we select one edge randomly from the second set and add to the new spanning tree, while we prevent generating circle by using union-find partition. The pseudo code of crossover operator is shown in Algorithm 2.\par

\begin{figure}[!ht]
\centering
\includegraphics[height=40mm]{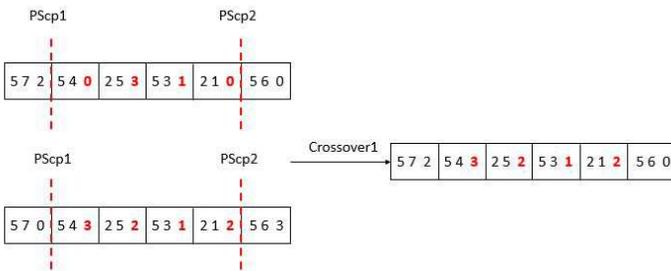}
\caption{Crossover operator with PS transformation.}\label{Fig.5}
\end{figure}

\begin{figure}[!ht]
\centering
\includegraphics[height=44mm]{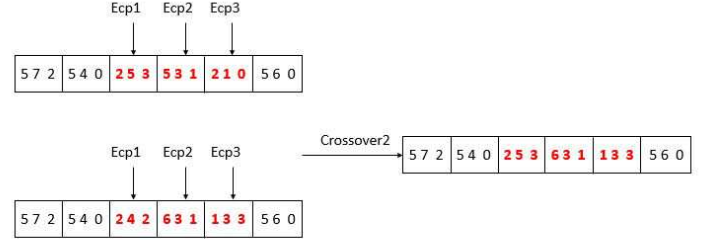}
\caption{Crossover operator with E transformation.}\label{Fig.5}
\end{figure}

\begin{algorithm}[t]
\caption{Crossover operator($p$, $q$)}
\begin{algorithmic}
\REQUIRE Particle  $p$ and $q$ \ENSURE New particle \STATE
$Initialize ~each ~pin's ~partition ~to ~singletons$ \STATE
$Sort\_edge(p,u);$//sort edge of $p$ according to the serial number
of the first endpoint $u$ \STATE $Sort\_edge(p,v);$//sort edge of
$p$ according to the serial number of the second endpoint $v$ \STATE
$Sort\_edge(q,u);$ \STATE $Sort\_edge(q,v);$ \STATE
$set1=Selecct\_same\_edge(p,q);$ \STATE
$set2=Select\_different\_edge(p,q);$ \STATE
$Union\_partition(u,v,set1);$//merge each edge of set1 \STATE $New~
particle=Generate\_edge(set1);$ \WHILE{$New ~particle ~is ~not~ a~
complete ~tree$} \STATE $L(u,v)=Random\_select_edge(set2);$
\IF{$Find\_set(u) \neq Find\_set(v)$} \STATE $ add~ L(u,v)~ to~ New~
particle;$ \STATE $Union\_partition(u,v,L);$ \ENDIF \ENDWHILE

\end{algorithmic}
\end{algorithm}
\subsection{Multi-Stage Transformation}
\indent  Considering the respective advantages of PS transformation and E transformation in different test circuits, we assume that these two transformations can be combined with different methods to find an optimal multi-stage transformation. Firstly, we divide the iterative process of the algorithm into the first half and the second half, namely two-stage transformation. As shown in Table \Rmnum{6}, we do away with all the combinations of transformations, including PS:PS, E: E, E:PS, PS: E. In Table \Rmnum{6}, for the combinations CM1 and CM2, the first half and the second half are using the same transformation strategy. CM1 of Table \Rmnum{6} is PS transformation, and CM2 of Table \Rmnum{6} is E transformation.\par

\textit{Property 8}
 E transformation strategy can enlarge the optimization space and PS transformation strategy can accelerate convergence in the later iteration.
 \par
\indent E:PS transformation strategy, known as CM2, refers to E transformation is adopted in the early stage of the iterative algorithm, and the latter only PS transformation is adopted. This two-stage transformation strategy CM2 can achieve the 9.91\% average improvements over the RSMT construction algorithm on wirelength, as shown in Table \Rmnum{6}. CM2 is the best combination for the two-stage transformation strategy in Table \Rmnum{6}.
\begin{table}[htbp]
  \centering
  \caption{Comparison among various combinations of the two-stage transformation strategy}
    \begin{tabular}{cccc}
    \hline
    \multirow{2}[2]{*}{Number} & First & Second & Imp/ \\
          & half  & half  & RSMT \\
    \hline
    CM1   & E     & E     & 9.76\% \\
    CM2   & E     & PS    & 9.91\% \\
    CM3   & PS    & E     & 9.84\% \\
    CM4   & PS    & PS    & 9.28\% \\
    \hline
    \end{tabular}%
  \label{tab:addlabel}%
\end{table}%

Secondly, we dived the the iterative process of the algorithm into the first stage, the second stage, and the third stage,, namely three-stage transformation.
\begin{table}[htbp]
  \centering
  \caption{Comparison among various combinations of the three-stage transformation strategy}
    \begin{tabular}{ccccc}
    \hline
    \multirow{2}[2]{*}{Number} & \multirow{2}[2]{*}{First} & \multirow{2}[2]{*}{Second} & \multirow{2}[2]{*}{Third} & Imp/ \\
          &       &       &       & RSMT \\
    \hline
    CM1   & E     & E     & E     & 9.73\% \\
    CM2   & E     & E     & PS    & 10.41\% \\
    CM3   & E     & PS    & E     & 10.45\% \\
    CM4   & E     & PS    & PS    & 10.45\% \\
    CM5   & PS    & E     & E     & 10.19\% \\
    CM6   & PS    & E     & PS    & 10.47\% \\
    CM7   & PS    & PS    & E     & 10.48\% \\
    CM8   & PS    & PS    & PS    & 10.22\% \\
    \hline
    \end{tabular}%
  \label{tab:addlabel}%
\end{table}%

PS:PS:E transformation strategy, known as CM7 can achieve the 10.48\% average improvements over the RSMT construction algorithm on wirelength, as shown in Table \Rmnum{7}. CM7 is the best combination for the three-stage transformation strategy in Table \Rmnum{7}.

Thirdly, we dived the the iterative process of the algorithm into the first stage, the second stage, the third stage, and the fourth stage, namely four-stage transformation.
\begin{table}[htbp]
  \centering
  \caption{Comparison among various combinations of the four-stage transformation strategy}
    \begin{tabular}{cccccc}
    \hline
    \multirow{2}[2]{*}{Number} & \multirow{2}[2]{*}{First} & \multirow{2}[2]{*}{Second} & \multirow{2}[2]{*}{Third} & \multirow{2}[2]{*}{Fourth} & Imp/ \\
          &       &       &       &       & RSMT \\
    \hline
    CM1   & E     & E     & E     & E     & 9.73\% \\
    CM2   & E     & E     & E     & PS    & 10.45\% \\
    CM3   & E     & E     & PS    & E     & 10.53\% \\
    CM4   & E     & E     & PS    & PS    & 10.49\% \\
    CM5   & E     & PS    & E     & E     & 10.43\% \\
    CM6   & E     & PS    & E     & PS    & 10.56\% \\
    CM7   & E     & PS    & PS    & E     & 10.44\% \\
    CM8   & E     & PS    & PS    & PS    & 10.45\% \\
    CM9   & PS    & E     & E     & E     & 10.14\% \\
    CM10  & PS    & E     & E     & PS    & 10.49\% \\
    CM11  & PS    & E     & PS    & E     & 10.51\% \\
    CM12  & PS    & E     & PS    & PS    & 10.47\% \\
    CM13  & PS    & PS    & E     & E     & 10.44\% \\
    CM14  & PS    & PS    & E     & PS    & 10.51\% \\
    CM15  & PS    & PS    & PS    & E     & 10.49\% \\
    CM16  & PS    & PS    & PS    & PS    & 10.24\% \\
    \hline
    \end{tabular}%
  \label{tab:addlabel}%
\end{table}%

E:PS:E:PS transformation strategy, known as CM6 can achieve the 10.54\% average improvements over the RSMT construction algorithm on wirelength, as shown in Table \Rmnum{8}. CM6 is the best combination for the four-stage transformation strategy in Table \Rmnum{8}.

Finally, we also test five-stage transformation, six-stage transformation, and then the best combination of these transformations is not better than the best one of four-stage transformation. Then it find that it is not divided into more stages, the better the convergence effect. So we select the best four-stage transformation for the proposed multi-stage transformation.
\subsection{Parameter Setting}
\textit{Property 9}
The setting of inertia weight affects the
balance between local search ability and global search ability of
particle.

As we can see from the velocity update formula, the first part
provides the flight impetus of particle in search space, and
represents the effect of previous velocity on the flight trajectory.
Thus inertia weight is a numerical value which indicates the extent
of such influence.

\textit{Property 10}
 Larger inertia weight will make the algorithm
has strong global search ability.

Property 9 and (1) show that inertia weight decides how much
previous velocity will be preserved. Thus a lager inertia weight can
strengthen the capability of searching the unreached area. It is
conductive to enhance the global search ability of the algorithm and
jump out of the local minima. A smaller inertia weight suggests that
the algorithm mainly search near the current solution. It is
conductive to enhance the local search ability and accelerate
convergence.

In the work [44], researchers presented a PSO algorithm based on
liner decreasing inertia weight. In order to ensure a stronger
global search, they employed a lager inertia weight early in the
program, and a smaller one in the later stages to guarantee the
local search. Simulation on four kinds of different benchmark
functions showed that such strategy of parameters actually improved
the performance of PSO.

\textit{Property 11:} Larger acceleration coefficients $c_1$ may
cause wandering in local scope. Larger acceleration coefficients
$c_2$ will make the algorithm prematurely converge on local optimal
solution.

Acceleration coefficients $c_1$ and $c_2$ are used in communicating
between particles. Ratnaweera et al. proposed a kind of strategy
which employ a lager $c_1$ and a smaller $c_2$ in the early phases
and the opposite in the later [45]. In this way, the algorithm will
guarantee detailed search in local scope, not have to directly move
to the position of global optimal in early phases., and speed up
convergence in the later stages. Similarly, experiment achieved
great results.

Based on the above analysis, we have tested 567 kinds of the
acceleration coefficients in Oliver30 TSP which is a minimization
problem [38]. Each experimental setting is conducted five runs and
each average is calculated. Consequently, $c_1$ = 0.82-0.5 and $c_2$
= 0.4-0.83 are considered as the optimal combination of parameter
settings. We adopt the idea of linear decline proposed by Shi and
Eberhart [44] and the optimal combination of parameter settings of
$c_1$ and $c_2$ to update the acceleration coefficients according to
(13) and (14). Besides, the other parameters in the proposed
algorithm are given as follows: $w$ decreases linearly from 0.95 to 0.4 according to (15) which is
similar to the acceleration coefficients.
\begin{equation}
c_1  = c_1 \_start\rlap{--} \rlap{--} \rlap{--}  - \frac{{c_1
\_start - c_1 \_end}}{{evaluations}} \times eval
\end{equation}
\begin{equation}
c_2  = c_2 \_start\rlap{--} \rlap{--} \rlap{--}  - \frac{{c_2
\_start - c_2 \_end}}{{evaluations}} \times eval
\end{equation}
\begin{equation}
w = w\_start\rlap{--} \rlap{--} \rlap{--}  - \frac{{w\_start -
w\_end_{} }}{{evaluations}} \times eval
\end{equation}
where $eval$ represents the current iteration and $evaluations$
represents the maximum number of iterations.

\subsection{Procedure of NPSO-MST-GO Algorithm}
The detail procedure of NPSO-MST-GO Algorithm can be summarized as follows:\par
\begin{itemize}
\item
Step 1: Load circuit net-list data and sort them in ascending order according to the size of coordinates.
\item
Step 2: Initialize the population size, maximum iterations, inertia weight, and acceleration factors. Then the initial population is randomly generated.
\item
Step 3: Calculate the fitness of each particle among the initial population according to (5). And the personal optimal solution of each particle is set as itself and select the particle with minimal fitness as the global optimal solution of the initial population.
\item
Step 4: The proposed efficient multi-stage transformation is used as the basic operation of particle update method.
\item
Step 5: Adjust the position and velocity of each particle according to (7)-(9), including  three steps: the retention of particle's previous velocity, learning the best direction of particles, and learning the optimal direction of the current population.
\item
Step 6: And then calculate the fitness value of each particle. If its fitness value is less than the personal optimal value, it is set as the personal optimal solution. If its fitness value is less than the global optimal value of the population, it is set as the global optimal solution.
\item
Step 7: Check the termination condition. If fulfilled, the run is terminated and the solution is obtained. Otherwise, go to Step5.
\end{itemize}
\subsection{Analysis and Convergence Proof of NPSO-MST-GO Algorithm}
\textit{Theorem 1}
The Markov chain of NPSO-MST-GO algorithmis finite and homogeneous.

$Proof$ Follow the similar method of  [46] and [41], we can proof that the Markov chain of NPSO-MST-GO is finite and homogeneous. The definition of Finite Markov Chain can be referred as [47].

\textit{Theorem 2}
Transition probability matrix of the Markov chain made up of NPSO-MST-GO is positive definite.

$Proof$ Follow the similar method of [47] and [41], we can proof that transition probability matrix of the Markov chain made up of NPSO-MST-GO is positive definite.\par
The limit theorem for Markov chain [47] is the basis for the convergence of the algorithm. The limit theorem explains that the long-term probability of Markov chain does not depend on its initial states.

\textit{Lemma 2}
If mutation probability , the algorithm is an ergodic irreducible Markov chain which has only one limited distribution and nothing to do with the initial distribution, moreover the probability at a random time and random state is greater than zero.

$Proof$  At the \textsl{t}-th time, the \textsl{j}-th state
probability distribution of population $X(t)$ is:
\begin{equation}
{P_j}(t) = \sum\limits_{j \in S} {{P_i}(1)P_{ij}^{(t)}} ,t = 1,2,
\cdots
\end{equation}

According to Theorem 2, we can get the formulation as following:
\begin{equation}
\begin{array}{l}
{P_j}(\infty ) = \mathop {\lim }\limits_{t \to \infty } (\sum\limits_{i \in S} {{P_i}(1)} P_{ij}^{(t)}) = \sum\limits_{i \in S} {{P_i}(1)P_{ij}^{(\infty )}}  > 0,\\
\forall j \in S
\end{array}
\end{equation}

\textit{Definition 9}
Suppose a stochastic variant ${Z_t} = \max \{ \\
f(x_k^{(t)}(i))|k = 1,2, \cdots ,N\} $ which represents individual
best fitness at the $t$-th step and $i$-th state of the population.
Then the algorithm converges to the global optimum, if and only if
\begin{equation}
\mathop {\lim }\limits_{t \to \infty } P\{ {Z_t} = {Z^*}\}  = 1
\end{equation}
where ${Z^*} = \max \{ f(x)|x \in S\} $ represents the global
optimum.

\textit{Theorem 3}
For any \textsl{i} and \textsl{j}, the time transiting of
an ergodic Markov chain from the \textsl{i-th} state to the
\textsl{j-th} state is limited.

\textit{Theorem 4}
NPSO-MST-GO algorithm can converge to the global
optimum.

$Proof$  Suppose that $i \in S$, ${Z_t} < {Z^*}$ and ${P_i}(t)$ is
the probability of  NPSO-MST-GO algorithm at \textsl{i}-th state
and the \textsl{t}-th step. Obviously $P\{ {Z_t} \ne {{\rm Z}^ * }\}
\ge {P_i}(t)$, hence we can know that $P\{ {Z_t} = {Z^*}\}  \le 1 -
{P_i}(t)$.

According to Lemma 2, the $i$-th state probability of the operator
in  NPSO-MST-GO algorithm is ${P_i}(\infty ) > 0$, then
\begin{equation}
\mathop {\lim }\limits_{t \to \infty } P\{ {Z_t} = {Z^*}\}  \le 1 -
{P_i}(\infty ) < 1
\end{equation}

Observe a new population such as $X_t^ +  = \{ {Z_t},{X_t}\} ,t \ge
1,{x_{ti}} \in S$ denoting the search space (which is a finite set
or a countable set), where ${Z_t}$, the same to that in Definition 9, represents individual best fitness in current population, ${X_t}$
denotes the population during the search. As it is easy to prove
that the group shift process $\{ X_t^ + ,t \ge 1\} $  is still a
homogeneous and ergodic Markov chain, we can know that
\begin{equation}
\begin{array}{l}
 P_j^ + (t) = \sum\limits_{i \in S} {P_i^ + (1)P_{ij}^ + (t)}  \\
 P_{ij}^ +  > 0{\rm{   }}(\forall i \in S,\forall j \in {S_0}) \\
 P_{ij}^ +  = 0{\rm{   }}(\forall i \in S,\forall j \notin {S_0}) \\
 \end{array}
\end{equation}
So

\begin{equation}
\begin{array}{l}
 {(P_{ij}^ + )^t} \to 0{\rm{   }}(t \to \infty ) \\
 P_j^ + (\infty ) \to 0{\rm{   }}(j \notin {S_0}) \\
 \mathop {\lim }\limits_{t \to \infty } P\{ {Z_t} = {Z^*}\}  = 1 \\
 \end{array}
\end{equation}
\section{Experiment results}
Two benchmark circuit suites, namely $GEO$ and $ISPD$ respectively, are used in this paper [48,49]. The circuit scales of $GEO$ and $ISPD$ are given in Table 10 and Table 13, respectively. It can be seen from Tables 10 and 13, the scale of $GEO$ including the pin number from 8 to 1000, while the scale of $ISPD$ including the total pin number from 44266 to 26900 and the number of nets from 11507 to 64227.

\subsection{Validation of multi-stage transformation}
Based on the best threshold parameter, the detail comparison among our proposed algorithm and the two similar algorithms [30,31] is given in Table \Rmnum{9}. From Table \Rmnum{9}, we can see that the constructed XSMT in this paper achieves the 0.98\% wirelength reduction compared with the constructed XSMT  [31] and the 10.56\% wirelength reduction compared with the constructed RSMT  [30]. It pointed out that the reduction ratio of XSMT length to RSMT length is generally 9.75$\pm $2.29\% [10]. And the average reduction rate of our algorithm is 10.56\%, which is within the optimum rate range.

Moreover, when the proposed algorithm is used to solve large-scale problems, such as Test 8,9,10, its optimization ability is becoming stronger and stronger. Therefore, the proposed algorithm is more advantageous and suitable for solving the large-scale routing problems.

\begin{table}[htbp]
  \centering
  \caption{Comparison results among various SMT construction algorithms}
    \begin{tabular}{rrrrrrr}
    \hline
    \multirow{2}[2]{*}{TEST} & \multirow{2}[2]{*}{PIN\#} & \multirow{2}[2]{*}{[30]} & \multirow{2}[2]{*}{[31]} & \multirow{2}[2]{*}{Ours} & \multicolumn{1}{l}{Imp
} & \multicolumn{1}{l}{Imp
} \\
          &       &       &       &       & \multicolumn{1}{l}{/[30]} & \multicolumn{1}{l}{/[31]} \\
    \hline
    1     & 8     & 17928 & 16918 & \multicolumn{1}{c}{16900} & 5.74\% & 0.11\% \\
    2     & 9     & 20478 & 18041 & \multicolumn{1}{c}{18023} & 11.99\% & 0.10\% \\
    3     & 10    & 21969 & 19696 & \multicolumn{1}{c}{19397} & 11.71\% & 1.52\% \\
    4     & 20    & 35675 & 32207 & \multicolumn{1}{c}{32038} & 10.20\% & 0.52\% \\
    5     & 50    & 53518 & 48020 & \multicolumn{1}{c}{47888} & 10.52\% & 0.27\% \\
    6     & 70    & 62633 & 56433 & \multicolumn{1}{c}{55771} & 10.96\% & 1.17\% \\
    7     & 100   & 75584 & 68855 & \multicolumn{1}{c}{68157} & 9.83\% & 1.01\% \\
    8     & 410   & 158206 & 141894 & \multicolumn{1}{c}{139505} & 11.82\% & 1.68\% \\
    9     & 200   & 170924 & 154825 & \multicolumn{1}{c}{152314} & 10.89\% & 1.62\% \\
    10    & 1000  & 246731 & 221090 & \multicolumn{1}{c}{217171} & 11.98\% & 1.77\% \\
    AVG   &       &       &       &       & 10.56\% & 0.98\% \\
    \hline
    \end{tabular}%
  \label{tab:addlabel}%
\end{table}%

\subsection{Validation of being extended to rectilinear architecture}
\label{sec:4.3}

We further apply the efficient multi-stage transformation proposed in this paper to be in the rectilinear architecture, that is, the RSMT construction based on our proposed  multi-stage transformation, and the experimental results are shown in Table \Rmnum{10}. In Table \Rmnum{10}, MRMST indicates that the proposed ulti-stage  transformation is involved, while RSMT indicates that the ulti-stage transformation is not involved. Experimental results have shown that the proposed ulti-stage  transformation designed in the rectilinear architecture routing can achieve 2.92\% of the average wirelength optimization than the method without the proposed ulti-stage transformation.

In summary, the proposed four-stage transformation can bring better wirelength optimization ability for both the rectilinear architecture and X-architecture, and it provides a good application prospect for designing the unified algorithm and then solving the routing problem based on different architectures. Finally, from Table \Rmnum{10}, the proposed four-stage transformation can not only further enhance the wirelentgh optimization of rectilinear architecture, but also has an ability to obtain accurate solutions.

\begin{table}[htbp]
  \centering
  \caption{The effectiveness of the RSMT construction algorithm based on the proposed four-stage transformation}
    \begin{tabular}{rrrrr}
    \hline
    \multirow{2}[2]{*}{TEST} & \multirow{2}[2]{*}{PIN\#} & \multirow{2}[2]{*}{RSMT} & \multirow{2}[2]{*}{MRMST} & \multicolumn{1}{l}{Imp} \\
          &       &       &       & \multicolumn{1}{l}{RSMT} \\
    \hline
    1     & 8     & 17928 & \multicolumn{1}{c}{17693} & \multicolumn{1}{c}{1.31\%} \\
    2     & 9     & 20478 & \multicolumn{1}{c}{19797} & \multicolumn{1}{c}{3.33\%} \\
    3     & 10    & 21969 & \multicolumn{1}{c}{21143} & \multicolumn{1}{c}{3.76\%} \\
    4     & 20    & 35675 & \multicolumn{1}{c}{34827} & \multicolumn{1}{c}{2.38\%} \\
    5     & 50    & 53518 & \multicolumn{1}{c}{51739} & \multicolumn{1}{c}{3.32\%} \\
    6     & 70    & 62633 & \multicolumn{1}{c}{60263} & \multicolumn{1}{c}{3.78\%} \\
    7     & 100   & 75584 & \multicolumn{1}{c}{74767} & \multicolumn{1}{c}{1.08\%} \\
    8     & 410   & 158206 & \multicolumn{1}{c}{153016} & \multicolumn{1}{c}{3.28\%} \\
    9     & 200   & 170924 & \multicolumn{1}{c}{165541} & \multicolumn{1}{c}{3.15\%} \\
    10    & 1000  & 246731 & \multicolumn{1}{c}{237287} & \multicolumn{1}{c}{3.83\%} \\
    AVG   &       &       &       & 2.92\% \\
    \hline
    \end{tabular}%
  \label{tab:addlabel}%
\end{table}%

\subsection{Validation of multiple topologies}
\label{sec:4.3}
For Test 3, four different topologies can be obtained by running this algorithm several times, as shown in Fig. 9. In fact, there are more than 4 kinds of topologies available with the same wirelength for Test 3, and only four of these topologies is listed in Fig. 9 due to limited space. The experimental results show that the algorithm can obtain many different topologies of RSMT and XSMT under the same or near optimal conditions. The ability to obtain multiple Steiner trees with different topologies can help to provide different topology options for congestion optimization in the global routing stage of VLSI.
\begin{figure*}[]
\centering
\subfigure[]{
\label{fig:subfig:a} 
\includegraphics[width=40mm,height=40mm]{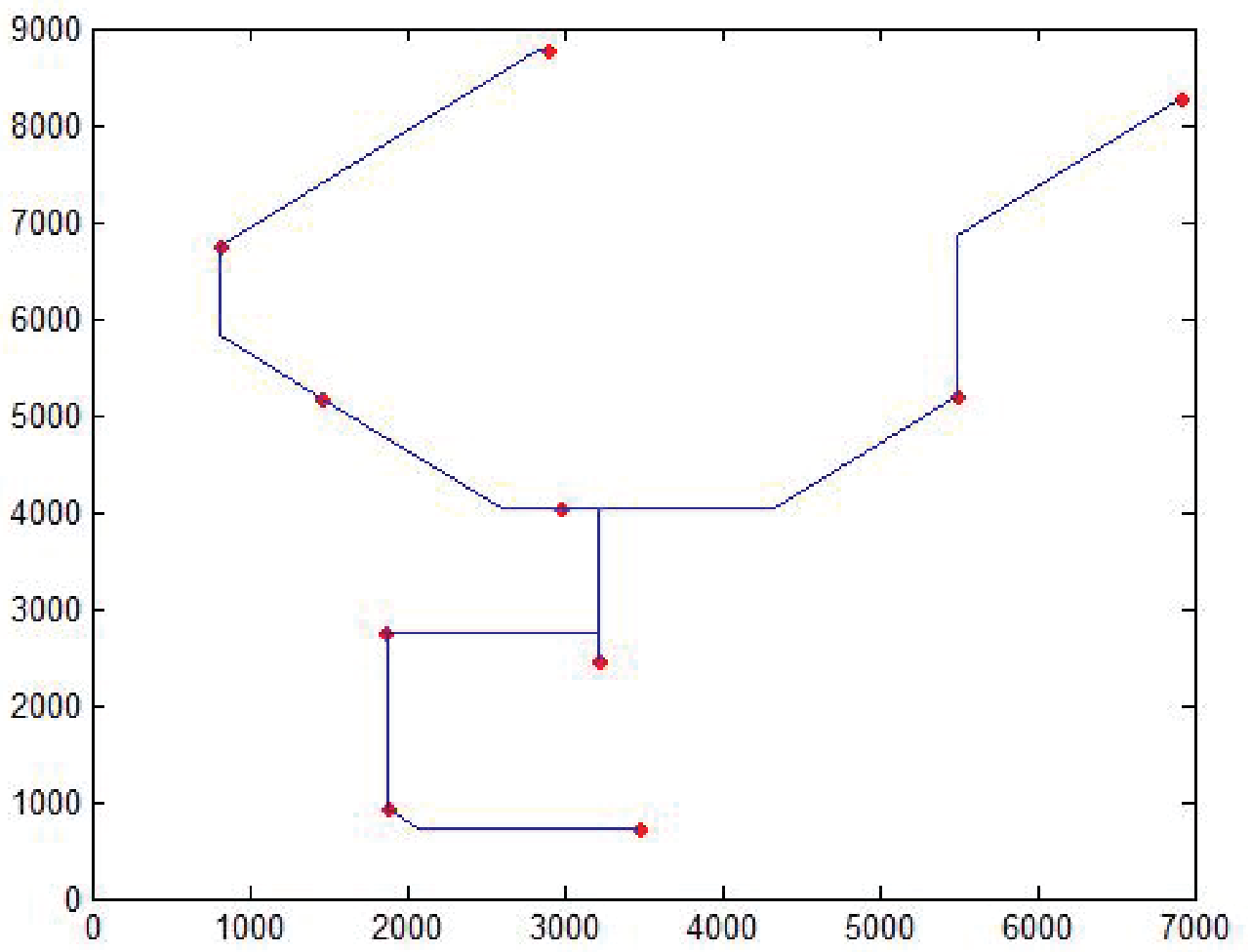}}
\subfigure[]{
\label{fig:subfig:b} 
\includegraphics[width=40mm,height=40mm]{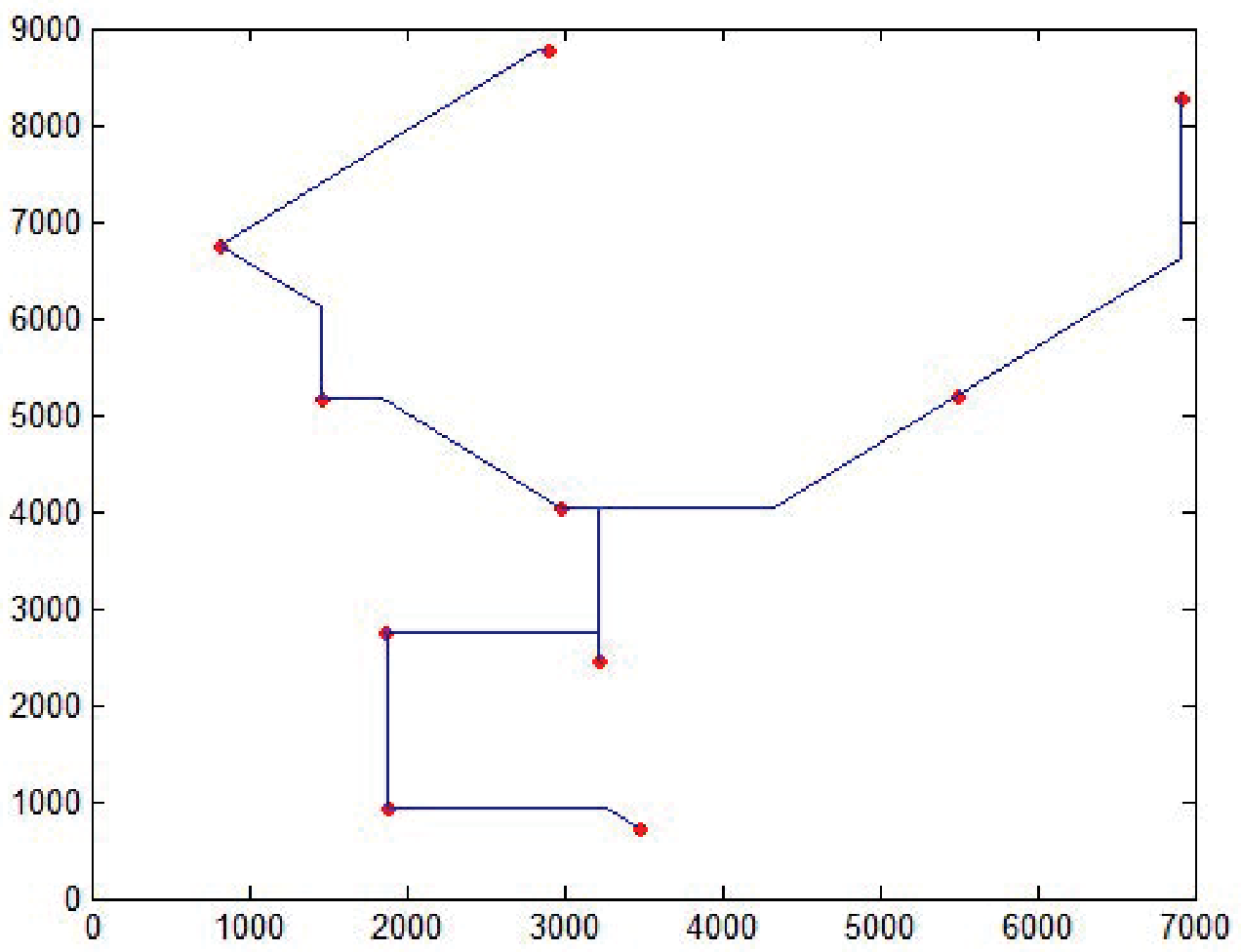}}
\subfigure[]{
\label{fig:subfig:c} 
\includegraphics[width=40mm,height=40mm]{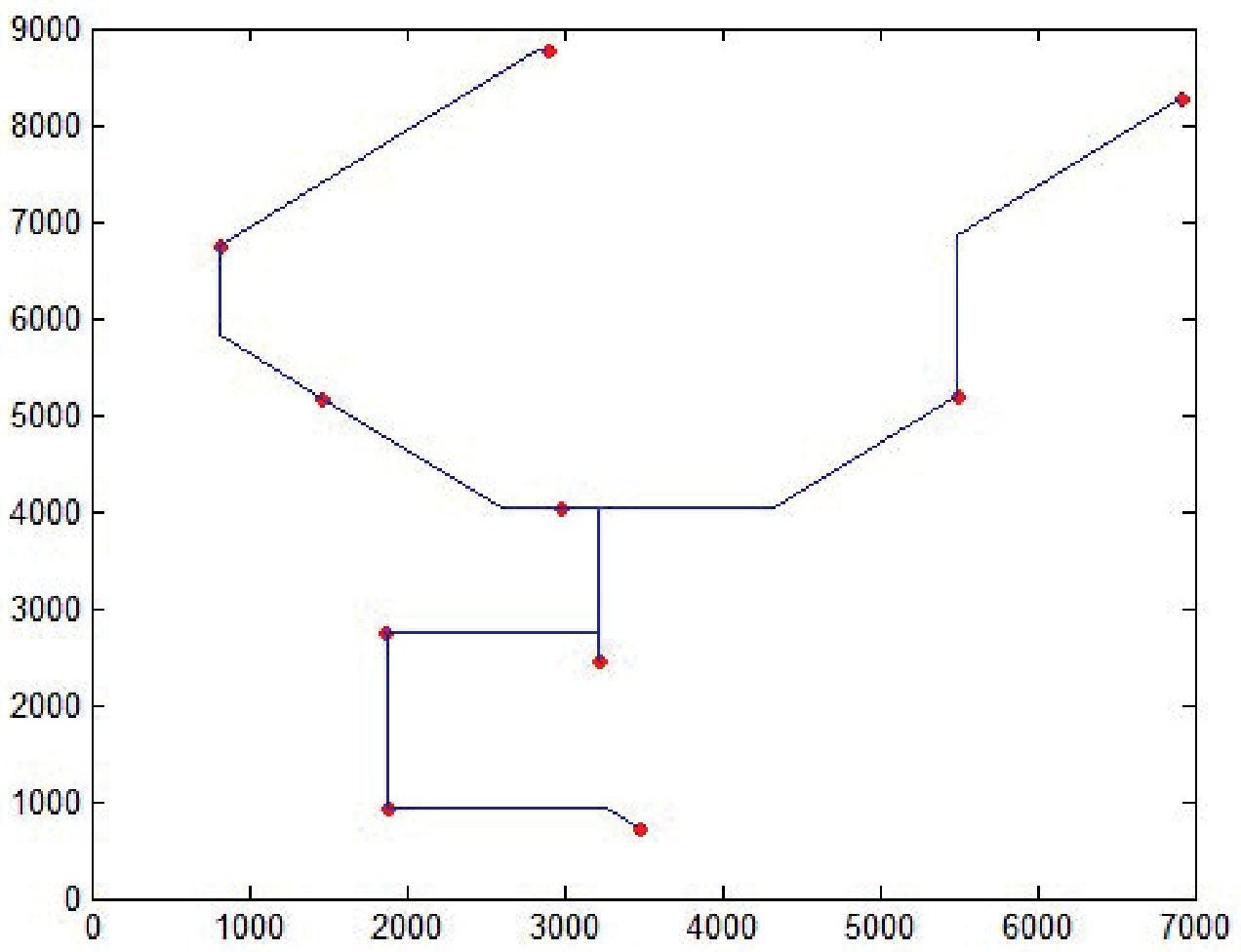}}
\subfigure[]{
\label{fig:subfig:d} 
\includegraphics[width=40mm,height=40mm]{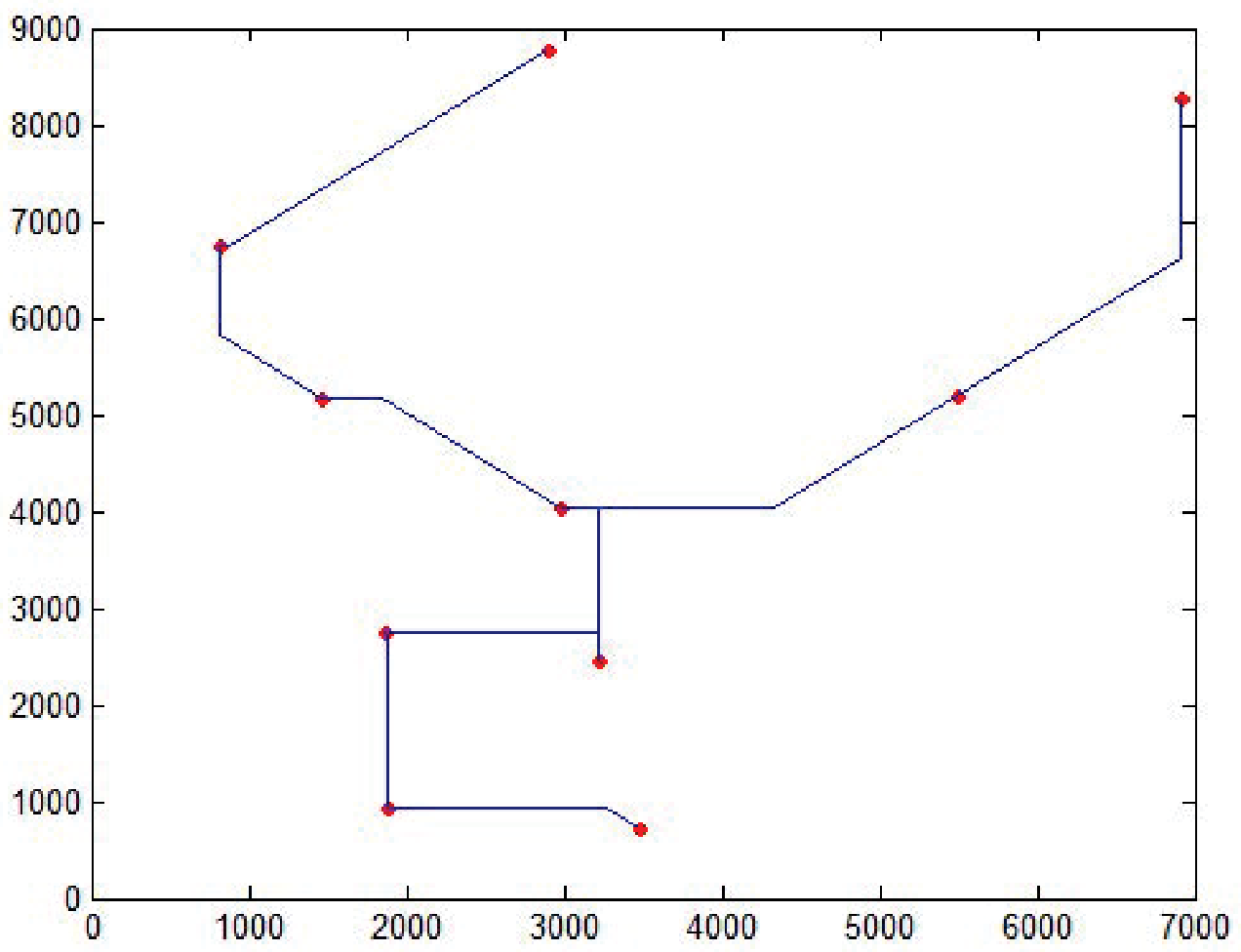}}
\caption{Four different topologies with the same wirelength for Test 3}
\label{fig:7:subfig} 
\end{figure*}

\subsection{Statistical results of several population based techniques }
To verify the superiority of the proposed method in terms of robustness, statistical results of the proposed method compared with two well established population based techniques are shown in Tables \Rmnum{11}-\Rmnum{12}. These two compared population based techniques are a differential evolution(DE) technique in [27], and an artificial bee colony optimization technique (ABC) in [28]. The experimental results of Table \Rmnum{11} and Table \Rmnum{12} are based on the benchmark circuits $ISPD$.

Each algorithm runs 20 times. In \Rmnum{11}, on the best value, our algorithms outperforms ABC and DE by 0.9\% and 1.2\%, respectively, while on the mean value, 0.9\% and 1.3\%, respectively. In Table \Rmnum{12}, on the standard deviation, our algorithm outperforms ABC and DE by 67.8\% and 45.7\%, respectively.

\begin{table*}[htbp]
  \centering
  \caption{Statistical results of several population based techniques in the circuits of Alpert (1998) (1)}
    \begin{tabular}{ccccccccccccc}
    \hline
    \multirow{3}[2]{*}{Circuit} & \multicolumn{6}{c}{Best Value}                & \multicolumn{6}{c}{Mean Value} \\
          & \multicolumn{3}{c}{Absolute Values} & \multicolumn{3}{c}{Normalised Values} & \multicolumn{3}{c}{Absolute Values} & \multicolumn{3}{c}{Normalised Values} \\
          & ABC   & DE    & Ours  & ABC   & DE    & Ours  & ABC   & DE    & Ours  & ABC   & DE    & Ours \\
    \hline
    ibm01 & 56730 & 56722 & \multicolumn{1}{r}{56096} & 1.011 & 1.011 & 1.000 & 56734 & 56726 & \multicolumn{1}{r}{56100} & 1.011 & 1.011 & 1.000 \\
    ibm02 & 156480 & 157287 & \multicolumn{1}{r}{155003} & 1.010 & 1.015 & 1.000 & 156495 & 157299 & \multicolumn{1}{r}{155011} & 1.010 & 1.015 & 1.000 \\
    ibm03 & 135217 & 135661 & \multicolumn{1}{r}{134051} & 1.009 & 1.012 & 1.000 & 135234 & 135671 & \multicolumn{1}{r}{134062} & 1.009 & 1.012 & 1.000 \\
    ibm04 & 151006 & 151539 & \multicolumn{1}{r}{149789} & 1.008 & 1.012 & 1.000 & 151024 & 151549 & \multicolumn{1}{r}{149794} & 1.008 & 1.012 & 1.000 \\
    ibm06 & 258951 & 260638 & \multicolumn{1}{r}{256789} & 1.008 & 1.015 & 1.000 & 258982 & 260666 & \multicolumn{1}{r}{256796} & 1.009 & 1.015 & 1.000 \\
    ibm07 & 338482 & 340042 & \multicolumn{1}{r}{335675} & 1.008 & 1.013 & 1.000 & 338512 & 340060 & \multicolumn{1}{r}{335679} & 1.008 & 1.013 & 1.000 \\
    ibm08 & 376283 & 378120 & \multicolumn{1}{r}{372406} & 1.010 & 1.015 & 1.000 & 376312 & 378132 & \multicolumn{1}{r}{372423} & 1.010 & 1.015 & 1.000 \\
    ibm09 & 385513 & 387186 & \multicolumn{1}{r}{382438} & 1.008 & 1.012 & 1.000 & 385549 & 387209 & \multicolumn{1}{r}{382472} & 1.008 & 1.012 & 1.000 \\
    ibm10 & 537828 & 539909 & \multicolumn{1}{r}{532837} & 1.009 & 1.013 & 1.000 & 537857 & 539940 & \multicolumn{1}{r}{532858} & 1.009 & 1.013 & 1.000 \\
    Average &       &       &       & 1.009 & 1.013 & 1.000 &       &       &       & 1.009 & 1.013 & 1.000 \\
    \hline
    \end{tabular}%
  \label{tab:addlabel}%
\end{table*}%

\begin{table}[htbp]
  \centering
  \caption{Statistical results of several population based techniques in the circuits of Alpert (1998) (2)}
    \begin{tabular}{ccccccc}
    \hline
    \multirow{3}[2]{*}{Circuit} & \multicolumn{6}{c}{Standard Deviation} \\
          & \multicolumn{3}{c}{Absolute Values} & \multicolumn{3}{c}{Normalised Values} \\
          & ABC   & DE    & Ours  & ABC   & DE    & Ours \\
    \hline
    ibm01 & 3.1   & 2.0   & \multicolumn{1}{r}{3.1 } & 1.006 & 0.651 & 1.000 \\
    ibm02 & 7.0   & 6.8   & \multicolumn{1}{r}{5.4 } & 1.293 & 1.245 & 1.000 \\
    ibm03 & 12.9  & 6.4   & \multicolumn{1}{r}{7.9 } & 1.629 & 0.812 & 1.000 \\
    ibm04 & 7.2   & 5.2   & \multicolumn{1}{r}{2.9 } & 2.451 & 1.776 & 1.000 \\
    ibm06 & 23.6  & 11.4  & \multicolumn{1}{r}{3.7 } & 6.319 & 3.045 & 1.000 \\
    ibm07 & 20.3  & 12.6  & \multicolumn{1}{r}{2.5 } & 8.283 & 5.127 & 1.000 \\
    ibm08 & 22.4  & 7.9   & \multicolumn{1}{r}{8.7 } & 2.576 & 0.908 & 1.000 \\
    ibm09 & 26.0  & 11.7  & \multicolumn{1}{r}{9.7 } & 2.671 & 1.205 & 1.000 \\
    ibm10 & 18.8  & 19.4  & \multicolumn{1}{r}{10.8 } & 1.732 & 1.786 & 1.000 \\
    Average &       &       &       & 3.107 & 1.840 & 1.000 \\
    \hline
    \end{tabular}%
  \label{tab:addlabel}%
\end{table}%

\subsection{Compared with recent published methods}
To verify the superiority of the proposed method, two recent published methods ([25], namely KNN in Table \Rmnum{13}, [26], namely SAT in Table \Rmnum{13}) are used to compared with our proposed algorithm and the results are shown in Tables \Rmnum{13}. In Table \Rmnum{13}, our algorithm reduces 10.00\% and 8.72\% than SAT and KNN on wirelength, respectively, while it speeds up 828.71X and 5.19X, respectively.
\begin{table*}[htbp]
  \centering
  \caption{Compared with SAT and KNN}
    \begin{tabular}{ccccccccccccc}
    \hline
    \multirow{3}[2]{*}{Bechmark} & \multirow{3}[2]{*}{\# Nets } & \multirow{3}[2]{*}{\# points} & \multicolumn{5}{c}{Wirelength}        & \multicolumn{5}{c}{Runtime(in min)} \\
          &       &       & \multirow{2}[1]{*}{SAT} & \multirow{2}[1]{*}{KNN} & \multirow{2}[1]{*}{Ours} & Imp   & Imp   & \multirow{2}[1]{*}{SAT} & \multirow{2}[1]{*}{KNN} & \multirow{2}[1]{*}{Ours} & Imp   & Imp \\
          &       &       &       &       &       & SAT   & KNN   &       &       &       & SAT   & KNN \\
    \hline
    ibm01 & 11507 & 44266 & 61005 & 61071 & \multicolumn{1}{r}{56100} & 8.04\% & 8.14\% & 152.63 & 3.77  & 1.07  & 142.65 & 3.52 \\
    ibm02 & 18429 & 78171 & 172518 & 167359 & \multicolumn{1}{r}{155011} & 10.15\% & 7.38\% & 1144.17 & 9.75  & 2.11  & 542.26 & 4.62 \\
    ibm03 & 21621 & 75710 & 150138 & 147982 & \multicolumn{1}{r}{134062} & 10.71\% & 9.41\% & 1817.97 & 10.75 & 1.78  & 1021.33 & 6.04 \\
    ibm04 & 26163 & 89591 & 164998 & 164838 & \multicolumn{1}{r}{149794} & 9.21\% & 9.13\% & 1993.45 & 11.55 & 2.01  & 991.77 & 5.75 \\
    ibm06 & 33354 & 124299 & 289705 & 280998 & \multicolumn{1}{r}{256796} & 11.36\% & 8.61\% & 3952.85 & 20.85 & 3.12  & 1266.94 & 6.68 \\
    ibm07 & 44394 & 164369 & 368015 & 368780 & \multicolumn{1}{r}{335679} & 8.79\% & 8.98\% & 4385.32 & 23.95 & 4.19  & 1046.61 & 5.72 \\
    ibm08 & 47944 & 198180 & 431879 & 413201 & \multicolumn{1}{r}{372423} & 13.77\% & 9.87\% & 5844.37 & 27.82 & 5.37  & 1088.34 & 5.18 \\
    ibm09 & 50393 & 187872 & 418382 & 417543 & \multicolumn{1}{r}{382472} & 8.58\% & 8.40\% & 3333.63 & 23.08 & 4.87  & 684.52 & 4.74 \\
    ibm10 & 64227 & 269000 & 588079 & 583102 & \multicolumn{1}{r}{532858} & 9.39\% & 8.62\% & 5250.10 & 34.50 & 7.79  & 673.95 & 4.43 \\
    Average &       &       &       &       &       & 10.00\% & 8.72\% &       &       &       & 828.71 & 5.19 \\
    \hline
    \end{tabular}%
  \label{tab:addlabel}%
\end{table*}%

\section{Conclusions}
A novel particle swarm optimizer with multi-stage transformation and genetic operators, namely NPSO-MST-GO, is proposed to construct two type of SMT, including non-Manhattan SMT and Manhattan SMT. Firstly, an effective edge-vertex encoding strategy with four types of pseudo-Steiner point choice is proposed for the non-Manhattan Steiner tree. It can be effectively extended to construct the Manhattan Steiner tree. Secondly, a multi-stage transformation strategy is proposed to both expand the algorithm search space and ensure the effective convergence. Thirdly, in order to bring uncertainty and diversity into the proposed PSO algorithm, four genetic operators are proposed. Finally, a series of simulation experiments are designed to illustrate the feasibility and effectiveness of the proposed strategies and can achieve better solution with less runtime. Delay is a very important optimization index and we will optimize the delay of chip design in future.
\ifCLASSOPTIONcaptionsoff
  \newpage
\fi

\begin{IEEEbiography}[{\includegraphics[width=0.8in,
height=1.25in,clip,keepaspectratio]{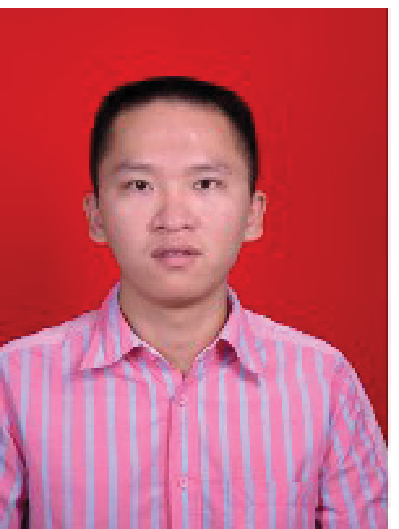}}]{Genggeng Liu} received the B.S. degree in Computer Science from Fuzhou University, Fuzhou, China, in 2009, and the
Ph.D degree in Applied Mathematics from Fuzhou University in 2015. He is currently an associate professor with the College of Mathematics and Computer Science at Fuzhou University.
His research interests include computational intelligence and very large scale integration physical design.
\end{IEEEbiography}


\begin{IEEEbiography}[{\includegraphics[width=0.9in, height=1.35in,clip,keepaspectratio]{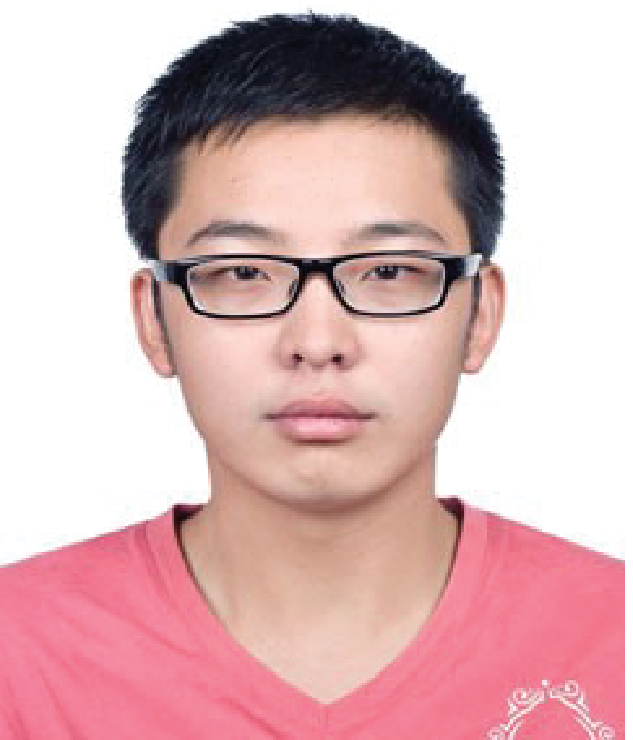}}]{Zhen Zhuang} is an undergraduate at the College of Mathematics and Computer Science from Fuzhou University. His research interests include computational intelligence and very large scale integration physical design.
\end{IEEEbiography}

\begin{IEEEbiography}[{\includegraphics[width=1.3in,
height=1.2in,clip,keepaspectratio]{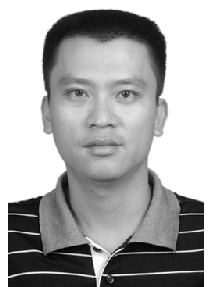}}]{Wenzhong Guo}
received the B.S. and M.S degrees in Computer Science from Fuzhou
University, Fuzhou, China, in 2000 and 2003, respectively, and the
Ph.D degree in Communication and Information System from Fuzhou
University in 2010. He is currently a full professor with the
College of Mathematics and Computer Science at Fuzhou University.
His research interests include mobile computing and evolutionary
computation. Currently, he leads the Network Computing and Intelligent Information Processing Lab, which
is a key Lab of Fujian Province, China.
\end{IEEEbiography}

\begin{IEEEbiography}[{\includegraphics[width=1.2in,
height=1.0in,clip,keepaspectratio]{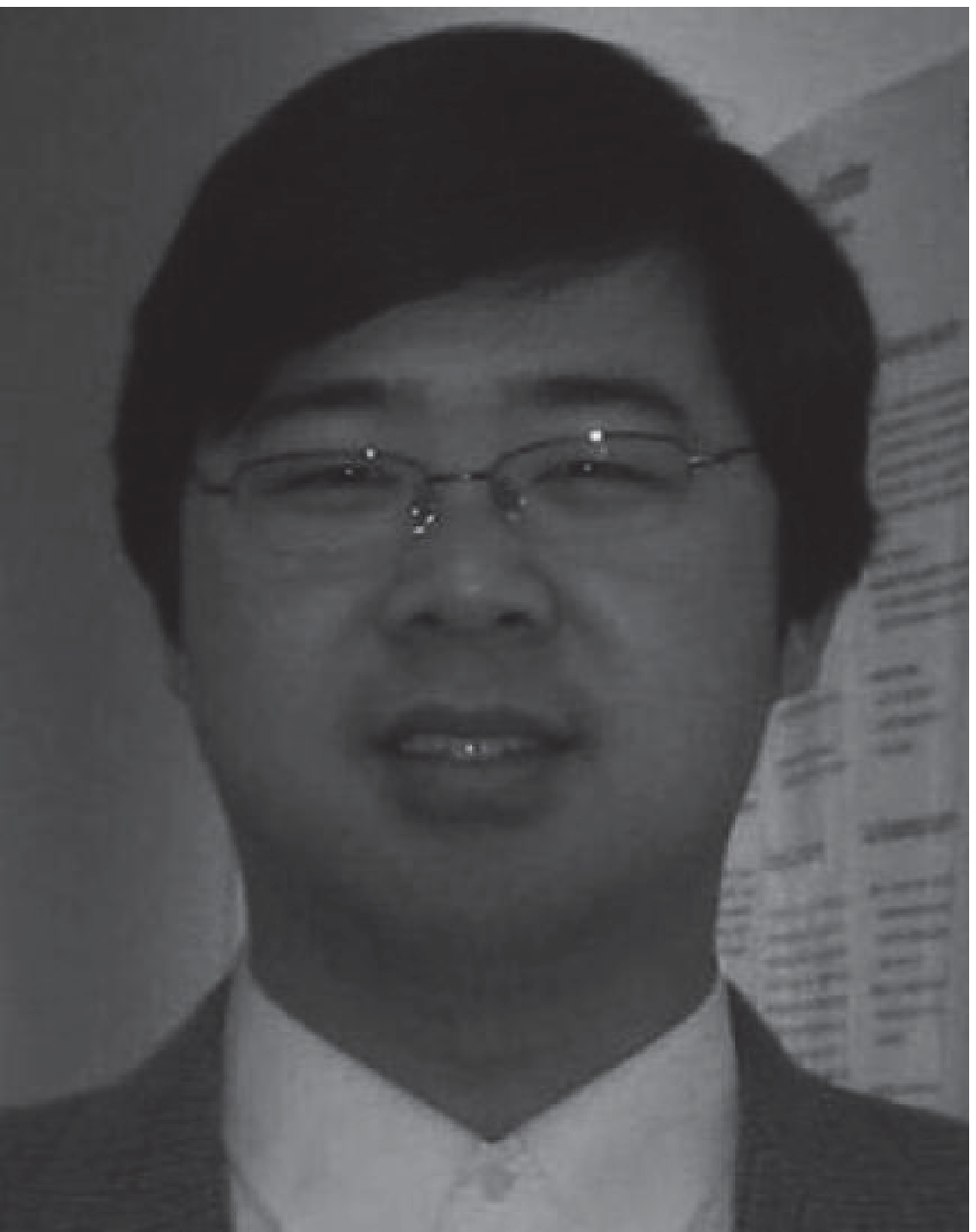}}]{Naixue Xiong}
received the B.E.
degree in computer science from the Hubei University
of Technology, Wuhan, China, in 2001,
the M.E. degree in computer science from Central
China Normal University, Wuhan, China, in
2004, and Ph.D. degrees in software engineering
from Wuhan University, Wuhan, China, in
2007, and in dependable networks from the
Japan Advanced Institute of Science and Technology,
Nomi, Japan, in 2008.
He is current a Full Professor at the Department
of Business and Computer Science, Southwestern Oklahoma State
University, Weatherford, OK, USA. His research interests include cloud
computing, security and dependability, parallel and distributed computing,
networks, and optimization theory.
Dr. Xiong serves as Editor-in-Chief, Associate Editor or Editor Member,
and Guest Editor for more than ten international journals including
as an Associate Editor of the IEEE TRANSACTIONS ON SYSTEMS, MAN
\& CYBERNETICS: SYSTEMS, and Editor-in-Chief of the Journal of Parallel
and Cloud Computing, the Sensor Journal, WINET, and MONET.
\end{IEEEbiography}

\begin{IEEEbiography}[{\includegraphics[width=1.2in,
height=1.1in,clip,keepaspectratio]{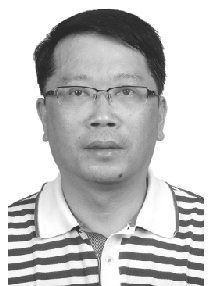}}]{Guolong Chen}
received the B.S. and M.S degrees in Computational Mathematics from
Fuzhou University, Fuzhou, China, in 1987 and 1992, respectively,
and the Ph.D degree in Computer Science from Xi'an Jiaotong
University, Xi'an, China, in 2002. He is a professor with the
College of Mathematics and Computer Science at Fuzhou University.
His research interests include computation intelligence, computer
networks, information security, etc.
\end{IEEEbiography}
\end{document}